\documentclass{article}

    \PassOptionsToPackage{numbers, compress}{natbib}
 \usepackage[preprint]{neurips_2026}

\usepackage[utf8]{inputenc} 
\usepackage[T1]{fontenc}    
\usepackage[hidelinks]{hyperref}      
\usepackage{url}            
\usepackage{booktabs}       
\usepackage{amsfonts}       
\usepackage{nicefrac}       
\usepackage{algorithm}
\usepackage{algpseudocode}

\usepackage{microtype}      
\usepackage{xcolor}         
\usepackage{amsmath,amsthm}
\usepackage{multirow}
\usepackage{subcaption}
\usepackage{makecell}
\usepackage{graphicx}

\title{Towards Generation-Efficient Uncertainty Estimation in Large Language Models}

\author{%
  Mingcheng Zhu \\
  University of Oxford\\
  \texttt{mingcheng.zhu@eng.ox.ac.uk} \\
  \And
  Yu Liu \\
  University of Oxford\\
  \texttt{yu.liu@eng.ox.ac.uk} \\
  \And
  Tingting Zhu \\
  University of Oxford\\
  \texttt{tingting.zhu@eng.ox.ac.uk} \\
}

\begin{document}

\maketitle

\begin{abstract}
Uncertainty estimation is important for deploying large language models (LLMs) in high-stakes applications such as healthcare and finance, where hallucinations can appear fluent and plausible while being factually incorrect, making it difficult for users to judge whether an output should be trusted. Existing methods typically require one or more full autoregressive generations to estimate uncertainty, which introduces substantial inference cost and often delays uncertainty assessment until generation completes. In this paper, we investigate whether effective uncertainty estimation can be achieved with partial generation or even input-only information. Specifically, we first develop a unified framework that formulates uncertainty estimation as an early estimation problem over the autoregressive generation process of LLMs. This framework organises existing and proposed estimators by the information they observe, ranging from multi-generation to input-only prediction, and clarifies the performance-cost trade-off underlying different uncertainty estimation methods. Building on this view, we study two largely underexplored low-cost settings: estimating uncertainty with part of the generation, and predicting uncertainty directly from the input prompt. We address these settings by proposing two estimators: Logit Magnitude, which uses top-$M$ logit evidence to estimate uncertainty from an early-stopped generation prefix, and MetaUE, which distils generation-based uncertainty into a lightweight input-only estimator trained with uncertainty scores. Extensive experiments on general and domain-specific benchmarks show that Logit Magnitude achieves strong performance across LLM families, scales, and architectures, including dense and MoE models, and partial generations of LLMs are often sufficient for effective uncertainty estimation. MetaUE further provides a competitive input-only approximation in several settings. These findings suggest that effective uncertainty estimation requires less generation than commonly assumed, enabling unreliable responses to be identified earlier. Code is available in the GitHub repository\footnote{\url{https://github.com/JasonZuu/Gen-Efficient_UE}}.
\end{abstract}

\section{Introduction}

\begin{figure}[h]
    \centering
    \includegraphics[width=0.9\linewidth]{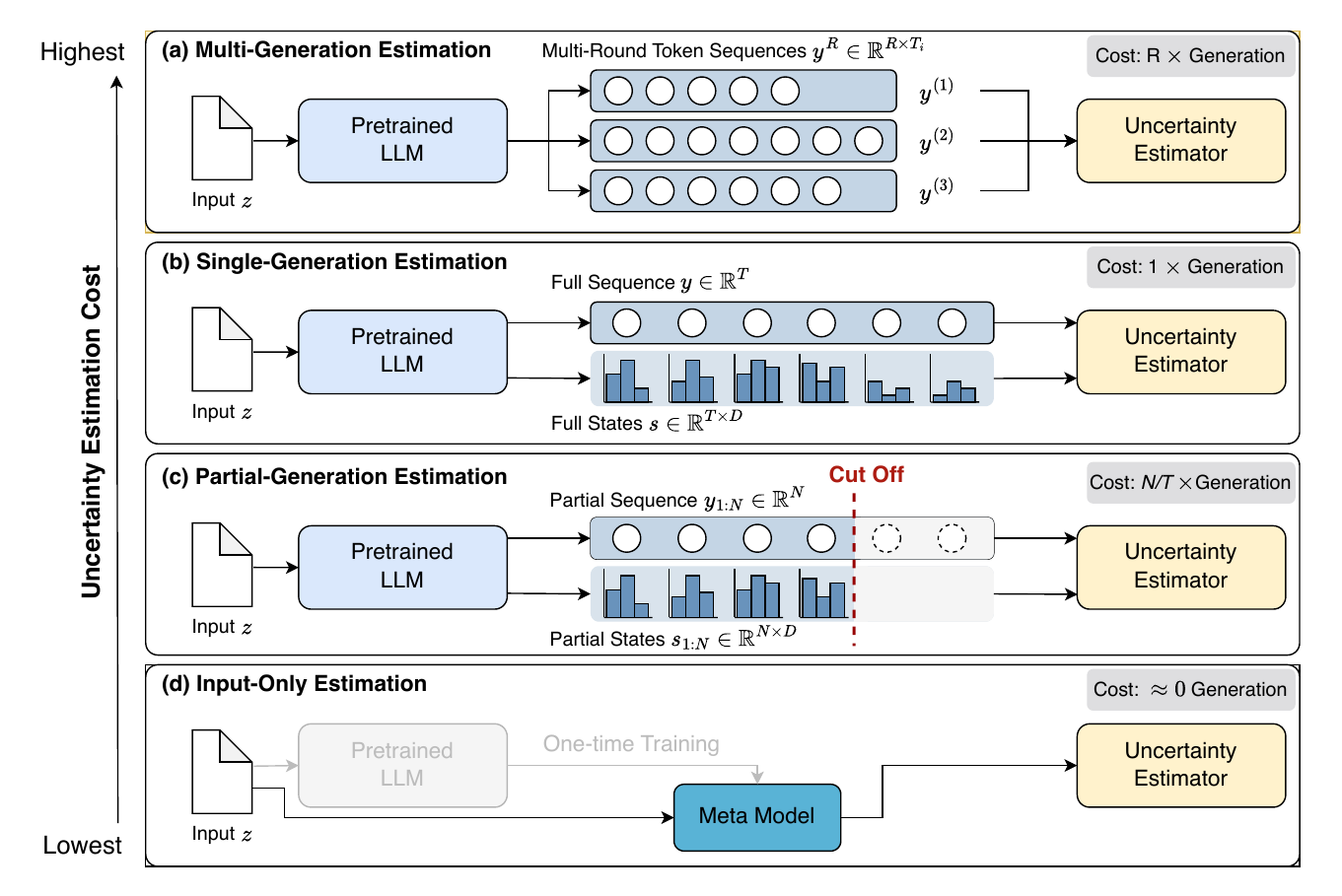}
    \caption{Taxonomy of LLM uncertainty estimation methods, ordered by estimation cost. The methods are defined by the portion of the generation sequence the estimator observes. \textbf{(a)} Estimating based on $R$ independent full generation sequences. \textbf{(b)} Observes one full generation sequence and the corresponding internal states. \textbf{(c)} Observes the first $N$ tokens of the generation sequence. \textbf{(d)} Skips autoregressive generation entirely, estimating uncertainty from the input prompt via a meta-model.}
    \label{fig:uq_methods}
\end{figure}

Large language models (LLMs) have demonstrated strong capabilities across diverse tasks, including open-ended generation~\cite{chang2025real}, question answering~\cite{chen2025knowledge}, and domain-specific reasoning~\cite{zhu2025medtpe}, motivating their deployment in broad real-world applications. However, LLMs can hallucinate during their autoregressive generation, producing fluent and plausible outputs that are factually incorrect or inconsistent with the preceding context~\cite{huang2025survey, vipulanandan2026semantic}. Such failures are especially concerning in high-stakes domains such as healthcare and finance, where overconfident errors can lead to misleading documentation~\cite{asgari2025framework}, inappropriate communication~\cite{mandal2025utilization}, or unreliable decision support~\cite{agweyu2026safety}. Therefore, reliable deployment requires not only generating useful responses but also estimating when those responses should not be trusted. 
To evaluate the reliability of LLMs, various uncertainty estimation methods have been proposed~\cite{huang2025survey}. These methods can be distinguished by how much of the autoregressive generation of the LLMs they observe, as illustrated in Figure~\ref{fig:uq_methods}. Specifically, multi-generation methods sample multiple generation sequences and estimate uncertainty from disagreement or inconsistency across generations~\cite{farquhar2024detecting, manakul2023selfcheckgpt} (Figure~\ref{fig:uq_methods}(a)). Although effective, their inference cost scales with the number of generations, leading to significantly higher latency and impractical for deployment~\cite{yang2025logu}. Single-generation methods reduce this overhead by estimating uncertainty from one full sequence, using signals such as sequence entropy~\cite{sharma2025think}, evidence score~\cite{ma2025estimating}, or hidden-state dynamics~\cite{bu2026sampling} (Figure~\ref{fig:uq_methods}(b)). However, such methods still require a full autoregressive generation before uncertainty can be assessed, creating an efficiency bottleneck for long-form generation and latency-sensitive applications, where token-by-token decoding often dominates inference latency~\cite{chen2025progressive}. These limitations motivate an important underexplored question: How much autoregressive generation is actually needed for effective LLM uncertainty estimation?

To address this question, we formalise LLM uncertainty estimation as an early estimation problem over the autoregressive generation process. As the model generates a response, each token and its associated internal status, such as probabilities, logits, or hidden representations, provide evidence about the reliability of the final answer~\cite{ma2025estimating}. This view naturally organises uncertainty estimators according to how much of the generation they observe, ranging from multi-generation and single-sequence estimation to partial-generation and input-only estimation. It also motivates our central hypothesis: reliable uncertainty estimation may not require observing the complete response if the most informative uncertainty signals are concentrated in an early or compact subset of the generation.

Motivated by this hypothesis, this paper focuses on two low-cost uncertainty estimation methods shown in Figure~\ref{fig:uq_methods}(c)-(d): partial-generation estimation and input-only prediction. For the partial-generation setting, we propose Logit Magnitude. At each generation step, an LLM produces logits, which are unnormalised internal features over candidate next tokens before the softmax operation. While softmax converts these scores into relative probabilities, the raw logits can retain information about the strength of the model's evidence~\cite{ma2025estimating}. Logit Magnitude extracts the magnitude of logits with L2 Norm as a token-level uncertainty score and aggregates the top-$M$ highest-scoring tokens in the observed prefix, allowing the sequence-level estimate to focus on the most informative uncertainty signals. For the input-only setting, we introduce MetaUE, a lightweight meta-model that predicts the Logit Magnitude uncertainty score directly from the prompt using a frozen pretrained encoder and a small trainable MLP head. In this way, the generation-based estimator provides pseudo-labels for MetaUE training, enabling uncertainty estimation before any response is generated.

Our contributions are threefold:
\textbf{(i)} We present a taxonomy of uncertainty estimation methods for autoregressive LLM generation, organising existing and proposed approaches according to the amount of generation information they observe, from multi-generation to input-only estimation.
\textbf{(ii)} We propose a generation-efficient uncertainty estimation framework that progresses from partial generation to zero generation. Our proposed Logit Magnitude with adaptive early-stop mechanism and top-$M$ aggregation estimates uncertainty from partial-generation and uses it as a generation-based pseudo-label to train MetaUE for input-only uncertainty estimation. 
\textbf{(iii)} We provide a systematic empirical study across general-domain and medical-domain datasets, multiple LLM families and model sizes, and MoE architectures, showing that informative partial-generation and input-only estimators can provide competitive uncertainty estimates at lower generation cost.

\section{Related Works}

\subsection{Multi-Generation Uncertainty Estimation} 

Multi-generation uncertainty estimation methods evaluate reliability by sampling several responses from the same input prompt and measuring disagreement or consistency across generations~\cite{manakul2023selfcheckgpt}. Self-checking methods treat inconsistency across sampled responses as evidence that a generation may be unsupported by the model's own distribution~\cite{manakul2023selfcheckgpt,ross2026textual}. Building on this intuition, sample-consistency methods use agreement among multiple sampled generations as a signal for calibrating LLM confidence~\cite{lyu2025calibrating,ganguly2025grammars}.
Semantic Entropy adapts this idea by grouping sampled responses into semantic equivalence clusters and quantifying uncertainty based on the differences among these clusters~\cite{farquhar2024detecting}. Another work improves this semantic view by encouraging semantically diverse generations, so that sampling can expose alternative meanings needed for reliable uncertainty estimation~\cite{aichberger2025improving, vashurin2025cocoa}. Recent work improves efficiency by replacing expensive sequence-level sampling with token-level sampling via low-rank weight perturbations~\cite{zhang2026tokur}. 
The main limitation of these methods is computational cost. Since each uncertainty estimate requires multiple full generations, the inference cost grows approximately linearly with the number of generations. Moreover, the multi-generation uncertainty score is for a set of sampled outputs rather than a single output. As a result, the score does not specify which candidate should be selected or whether any individual candidate is reliable enough to use~\cite{agrawal2025uncertainty}.

\subsection{Single-Generation Uncertainty Estimation } 

Single-generation methods reduce the cost of generation by estimating uncertainty from one full generation. Among them, probability-based approaches use the predictive distribution over generated tokens as a confidence signal for the final answer. For example, sequence probabilities are used to compute length-normalised metrics, which aggregate individual token likelihoods into a single measure of the model's overall certainty~\cite{sharma2025think, aichberger2026rethinking, bakman2026uncertainty}. Since free-text answers contain semantically redundant tokens, the relevance-aware or modality-aware method further reweights token-level uncertainty according to each token's contribution to the final answer~\cite{duan2024shifting, ju2025exploiting}. Other approaches exploit richer model-internal signals. Evidence-based uncertainty estimation treats logits as token-level evidence and maps them into Dirichlet-inspired scores~\cite{ma2025estimating}. Hidden-state approaches instead estimate reliability from representation dynamics across layers and generation steps, using signals from the hidden state of LLMs~\cite{bu2026sampling, zhou2026can}. Training-based variants include learning uncertainty from internal states to output its uncertainty during the autoregressive generation~\cite{vazhentsev2025unconditional}. Recent work improves the reliability of the confidence provided in the generation of LLMs through prompt design or post-training~\cite{zhou2025steerconf, damani2026beyond}. 
However, single-generation estimation still requires the model to generate a full sequence before the uncertainty becomes available. This is inefficient when the answer is long, and it is undesirable when the goal is to detect unreliable outputs as early as possible~\cite{liu2025answer}. In addition, many sequence-level estimators aggregate information across all generated tokens, which can dilute the effect of a small number of correctness-relevant tokens~\cite{zhou2026can}. Rather than assuming the full generation sequence is required, we examine whether uncertainty can be reliably estimated from a short generation prefix and a sparse subset of high-uncertainty tokens, and whether these signals can be predicted directly from the input.

\section{Problem Formulation}
\label{sec:problem formulation}

\textbf{Autoregressive Generation\quad} 
Let $\theta$ denote a pretrained conditional generative model, $\mathbf{z}$ be an input prompt, and $\mathbf{Y}=(y_1,\ldots, y_T)$ the generated response, where $T$ is the stopping length induced by an end-of-sequence token or a maximum generating budget. The conditional generation distribution
\begin{equation}
    p_{\theta}(\mathbf{y}\mid \mathbf{z})
    =
    \prod_{t=1}^{T}
    p_{\theta}(y_t \mid \mathbf{z}, \mathbf{y}_{<t}) .
    \label{eq:generation_distribution}
\end{equation}
At each generation step $t$, before sampling the next token, the model produces a vector of logits
$\mathbf{\ell}_t \in \mathbb{R}^{|\mathcal{V}|}$ over the vocabulary $\mathcal{V}$. Each logit is an unnormalised score assigned to candidate token $v$, and the next-token probability distribution $p_{\theta}(Y=\mathbf{y}\mid \mathbf{z}, \mathbf{y}_{<t})$ is obtained by applying the softmax function. In this work, the observable internal state $\mathbf{s}_t$ is the logit vector $\mathbf{\ell}_t$, from which we construct the token-level uncertainty score. We will use $\mathbf{s}_t$ to stand for the logit vector in the following content.

\textbf{Uncertainty Estimation\quad}
For a generated response $\mathbf{Y}$, let $\mathbf{y}^{\star}$ denote the reference answer or task-specific evaluation target. We define the unreliability variable as
\begin{equation}
    U = 1 - C=
    1 - f(\mathbf{y}, \mathbf{y}^{\star}),
    \qquad
    f(\mathbf{y}, \mathbf{y}^{\star}) \in [0,1],
    \label{eq:unreliability_variable}
\end{equation}
where $f$ is a task-specific correctness function, such as exact match, token-level similarity, semantic similarity, or another evaluation metric. Larger values of $f(\mathbf{y}, \mathbf{y}^{\star})$ indicate more reliable generations, while larger values of $U$ indicate greater unreliability. 


Let $\mathcal{I}$ denote the information available to the estimator, such as the input prompt, the logits, or the multi-generation. An uncertainty estimator is a scoring function
$u_{\psi}(\mathcal{I})$, where a larger value indicates that the response is predicted to be less reliable. Ideally, this score should approximate the expected unreliability conditioned on the available information:
\begin{equation}
    u^{\star}(\mathcal{I})
    =
    \mathbb{E}[U \mid \mathcal{I}].
    \label{eq:optimal_uncertainty_score}
\end{equation}
Let $u^{\star}(\mathcal{I})$ is the optimal estimator of $U$ under the mean-squared error objective:
\begin{equation}
    u^{\star}
    =
    \arg\min_{u}
    \mathbb{E}\!\left[
    \left(
    U-u(\mathcal{I})
    \right)^2
    \right].
    \label{eq:uncertainty_mse_objective}
\end{equation}
When correctness is binary, $u^{\star}(\mathcal{I})$ becomes the conditional probability that the response is incorrect. In practice, we evaluate uncertainty scores by whether they assign higher values to incorrect generations than to correct ones, since this ranking determines the usefulness for uncertainty detection.

\textbf{Observed Information\quad} 
We use the filtration~\cite{williams1991probability} to define the information available during generation
\begin{equation}
\mathcal{F}_0=\sigma(\mathbf{z}), \qquad
\mathcal{F}_t=\sigma(\mathbf{z},Y_{1:t},s_{1:t})\ \text{with}\  1\leq t\leq T, \qquad
\mathcal{F}^{(R)}_T=
\sigma\!\left(\mathbf{z},\{Y^{(r)}_{1:T_r},s^{(r)}_{1:T_r}\}_{r=1}^{R}\right),
\label{eq:generation_filtration}
\end{equation}
where $\mathcal{F}_0$ contains only the input prompt $\mathbf{z}$, $\mathcal{F}_t$ contains the information available after observing the first $t$ generated tokens $Y_{1:t}$ and their associated states $s_{1:t}$, and $\mathcal{F}^{(R)}_T$ contains the information from $R$ independently sampled full generations. 
This filtration provides a unified way to describe uncertainty estimators according to the amount of generation information they are allowed to observe.

\textbf{Estimation Cost\quad} In this formulation, the existing and proposed uncertainty estimation methods can be organised into four settings, as illustrated in Figure~\ref{fig:uq_methods}. Multi-generation estimation observes $\mathcal{F}^{(R)}_T$ by sampling multiple generations and estimating uncertainty from disagreement or inconsistency across responses. Its generation cost is $\mathcal{O}(R\cdot\mathcal{C}_{\mathrm{gen}})$, where $\mathcal{C}_{\mathrm{gen}}$ denotes the cost of one full autoregressive generation. Single-generation estimation observes $\mathcal{F}_T$, using one full generation and its associated states, with cost $\mathcal{O}(\mathcal{C}_{\mathrm{gen}})$. Partial-generation estimation observes $\mathcal{F}_{\tau}$, where $\tau\leq T$ is an adaptive stopping time determined by the information available up to that point. Its expected generation cost is approximately $\mathcal{O}((\tau/T)\cdot\mathcal{C}_{\mathrm{gen}})$. Finally, the input-only estimation observes only $\mathcal{F}_0$ and predicts the uncertainty directly from the prompt before the autoregressive generation begins. It therefore replaces the autoregressive generation with a single forward pass of the meta model.

\section{Methodology}
\label{sec:method}
\begin{figure}[ht]
\vspace{-0.5cm}
    \centering
    \includegraphics[width=0.9\linewidth]{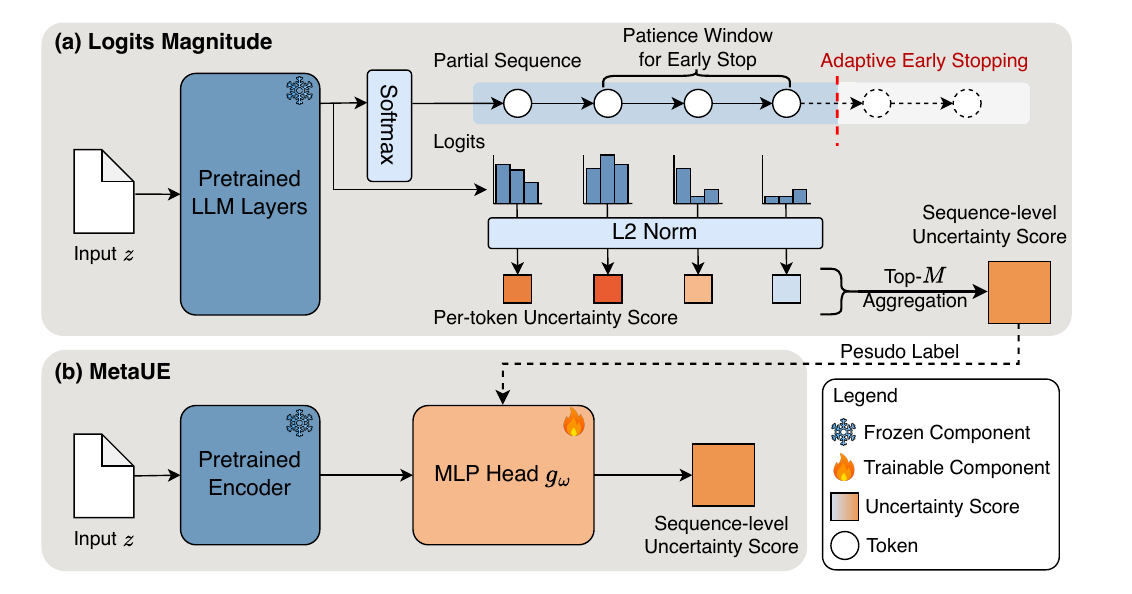}
    \vspace{-5px}
    \caption{
    Overview of the proposed generation-efficient uncertainty estimation framework.
    \textbf{(a) Logit Magnitude:} Given an input prompt, the frozen LLM generates a partial sequence and exposes the logits at each generation step. The logits are converted into token-level uncertainty scores using $L_2$-norm of logits. An adaptive early-stopping rule monitors whether the top-scoring tokens remain unchanged within a patience window, and the retained token-level scores are aggregated with top-$M$ aggregation to produce the sequence-level uncertainty score.
    \textbf{(b) MetaUE:} A frozen pretrained encoder maps the input prompt into a representation, and a lightweight trainable MLP head predicts the sequence-level uncertainty score before generation. MetaUE is trained using the sequence-level Logit Magnitude scores as the pseudo-labels.
    }
    \label{fig:our_method}
    \vspace{-0.2cm}
\end{figure}

\subsection{Logit Magnitude for Partial-Generation Uncertainty Estimation}
\label{sec:early_stop_relu_logits}

Probability-based uncertainty scores, such as sequence entropy~\cite{sharma2025think}, normalise logits into relative probabilities with softmax and may discard the absolute strength of model evidence. Recent work suggests that raw logits retain useful information about evidence strength and model reliability~\cite{ma2025estimating, zhou2026can}. We therefore use the positive Logit Magnitude as a token-level signal for uncertainty estimation. Intuitively, tokens with strong positive evidence can have a large influence on the final response, so their reliability is especially important for sequence-level uncertainty. Formally, for generation step $t$, let $\ell^{(K)}_t=(\ell^{(1)}_t,\ldots,\ell^{(K)}_t)$ denote the top-$K$ logits emitted by the model. We therefore compute the magnitude using positive logits and define the token-level score for the $t$-th generated token as
\begin{equation}
    u_t
    =
    \sum_{k=1}^{K}
    \max\!\sqrt{\left(\ell^{(k)}_t,0\right)^2}.
    \label{eq:token_level_logit_magnitude}
\end{equation}

We use an adaptive stopping strategy based on the token-level uncertainty scores. For each prompt $\mathbf{z}_i$, generation proceeds token by token. At generation step $t$, let $\mathcal{H}_t$ denote the indices of the current $\min(M,t)$ largest token-level scores among $u_{1:t}=(u_1,\ldots,u_t)$. We maintain a patience counter $n_t$, which records the number of consecutive generation steps since $\mathcal{H}_t$ was last updated. If the newly generated token has a score $u_t$ that enters the current top-$M$ set, we update $\mathcal{H}_t$ and reset $n_t=0$. Otherwise, $\mathcal{H}_t$ remains unchanged and the patience counter increases. Generation stops at $\tau_i=t$ if the LLM generates an end-of-sequence token, or if the top-$M$ set has remained unchanged for $W$ consecutive steps. After stopping, we compute the sequence-level uncertainty score using the final top-$M$ set $\mathcal{H}_{\tau_i}$. Formally,
\begin{equation}
    U_{\tau_i}
    =
    \frac{1}{|\mathcal{H}_{\tau_i}|}
    \sum_{t\in \mathcal{H}_{\tau_i}} u_t .
    \label{eq:sequence_level_logit_magnitude}
\end{equation}

To reduce sensitivity to the raw scale of $U_{\tau_i}$, we apply min-max normalisation over the training set and map the score to $[0,1]$. The detailed algorithm is illustrated in the Appendix~\ref{app:algorithm_logit_magnitude}.

The motivation for early stopping is that the gap between single-generation and partial-generation uncertainty estimation depends on the amount of correctness-relevant information remaining after the stopping time, rather than simply on the number of unobserved tokens. In Appendix~\ref{app:early_stop_martingale}, we prove that when the unobserved tail has small conditional update energy, the estimation error introduced by stopping early admits an upper bound. The patience rule provides an empirical realisation of this condition: if no new token enters the current top-$M$ set for $W$ consecutive steps, the recent updates to the sequence-level uncertainty estimate are likely to be small, suggesting that further generation is unlikely to substantially change the estimate.

\subsection{Meta Model for Input-Only Uncertainty Estimation}
\label{sec:MetaUE}

Despite Logit Magnitude reducing the amount of generation required for uncertainty estimation, it still depends on $\tau$-step autoregressive generation. To further reduce the cost of autoregressive generation, we introduce MetaUE, an input-only meta-model that predicts uncertainty directly from the prompt. MetaUE consists of a frozen pretrained text encoder $\phi$ and a trainable two-layer MLP head $g_{\omega}$. Given an input prompt $\mathbf{z}$, the MetaUE is defined as
\begin{equation}
    f_{\omega}(\mathbf{z})
    =
    g_{\omega}(\phi(\mathbf{z})),
    \label{eq:MetaUE}
\end{equation}
 where $f_{\omega}(\mathbf{z})$ is the predicted uncertainty score. Since $\phi$ is frozen and only $g_{\omega}$ is trained, MetaUE introduces only a few trainable parameters $\omega$, and avoids generation cost at inference.  
To train MetaUE without ground-truth correctness labels, we use the early-stopped Logit Magnitude score as a label-free distillation target. 
This allows MetaUE to distil generation-based uncertainty into an input-only estimator without requiring task-specific annotations. Specifically, MetaUE is trained by minimising the mean squared error between the score predicted from the input prompt and the uncertainty score produced by the early-stopped Logit Magnitude estimator. To avoid reliance on ground-truth correctness labels, which can be costly to obtain~\cite{zhen2025enhancing}, we use the proposed uncertainty score as a pseudo-label. Formally, for each training prompt $\mathbf{z}_i$, we compute the Logit Magnitude score $U_{\tau_i}$, which is used to train MetaUE with an MSE loss:
\begin{equation}
    \mathcal{J}(\omega)
    =
    \frac{1}{n}
    \sum_{i=1}^{n}
    \left(
    f_{\omega}(\mathbf{z}_i)-U_{\tau_i}
    \right)^2.
    \label{eq:MetaUE_mse_loss}
\end{equation}
The detailed training process of the MetaUE is illustrated in the Appendix~\ref{app:algorithm_metaue}.

\section{Experiments}
\label{sec:result}

\subsection{Experiment Setup}
\label{subsec:experiment setup}

\textbf{Datasets\quad}
We evaluate the proposed framework on three free-text question-answering datasets spanning general and medical domains: COQA~\cite{reddy2019coqa}, NewsQA~\cite{trischler2017newsqa}, and emrQA~\cite{pampari2018emrqa}. COQA and NewsQA are widely used benchmarks for evaluating uncertainty estimation in general-domain generation~\cite{kuhn2023semantic,bu2026sampling}. We additionally use emrQA to assess reliability in a high-stakes clinical setting based on electronic medical records. Further dataset details are provided in Appendix~\ref{app:dataset_details}.

\textbf{LLM Evaluation\quad}
Three LLM families, Qwen3.5~\cite{team2026qwen3}, Gemma4~\cite{team2024gemma}, and Llama3~\cite{grattafiori2024llama} are evaluated. Each model uses identical sampling parameters for all uncertainty estimators.

\textbf{Baseline Methods\quad} 
For multi-generation estimation baselines, we include the most representative methods, Semantic Entropy~\cite{farquhar2024detecting} and Self-Consistency~\cite{manakul2023selfcheckgpt}. For single-generation baselines, we include the representative methods entropy~\cite{sharma2025think} and LogTokU~\cite{ma2025estimating}, and the latest methods hidden-state score~\cite{bu2026sampling} and self-certainty~\cite{kang2025scalable}. 

\textbf{Evaluation Metrics\quad}
To evaluate the performance and efficiency of the uncertainty estimation methods, we report AUROC, AURAC, balanced accuracy (Bal. Acc), and the averaged number of tokens used for uncertainty estimation (N-tok). Details of metrics are provided in Appendix~\ref{app:evaluation_metrics}.

\subsection{Comparative Performance of Logit Magnitude and MetaUE}

This experiment compares uncertainty estimation methods across datasets and LLMs, assessing whether low-cost estimators can approach the performance of more expensive generation-based baselines. In particular, we compare full-generation Logit Magnitude, Logit Magnitude with adaptive early-stop, and the input-only MetaUE model against multi-generation and single-generation uncertainty estimation baselines. 
As shown in Table~\ref{table:baselines_comparison}, Logit Magnitude is an effective uncertainty estimator across both general-domain and medical-domain tasks. In the full-generation setting $(\tau=T)$, it achieves the best or highly competitive performance on most LLM-dataset combinations, indicating that the raw Logit Magnitude provides a strong signal for distinguishing reliable and unreliable generations. The early-stopped variant preserves most of this performance while using fewer generated tokens. For example, on Qwen3.5-4B, Logit Magnitude achieves an AUROC of $0.791$ on COQA compared with $0.793$ for full generation, and $0.750$ on NewsQA compared with $0.751$, while reducing the average token budget from $19.4$ to $16.3$ and from $15.7$ to $11.7$, respectively. Similar behaviour is observed on Gemma4-4B. These results suggest that reliable uncertainty estimation does not always require observing the full sequence. Finally, MetaUE provides a zero-generation alternative, achieving performance comparable to several generation-based baselines, showing that useful uncertainty information can also be distilled into an input-only estimator. 

\begin{table*}[h!]
\centering
\footnotesize
\caption{
Assessment of uncertainty estimation methods. For each LLM, methods are grouped into multi-generation baselines, single-generation baselines, and our proposed methods. \textbf{Bold} values indicate the best performance within each LLM-dataset combination, and \underline{underlined} values indicate the second-best performance within each combination.
}
\setlength{\tabcolsep}{2.0pt}
\renewcommand{\arraystretch}{1.08}
\label{table:baselines_comparison}
\resizebox{\textwidth}{!}{
\begin{tabular}{cc|cccc|cccc|cccc}
\toprule
\multirow{2}{*}{\textbf{LLM}}
& \multirow{2}{*}{\begin{tabular}{c}\textbf{UE}\\\textbf{Method}\end{tabular}}
& \multicolumn{4}{c|}{\textbf{COQA}}
& \multicolumn{4}{c|}{\textbf{NewsQA}}
& \multicolumn{4}{c}{\textbf{emrQA}} \\
\cmidrule(lr){3-6} \cmidrule(lr){7-10} \cmidrule(lr){11-14}
& & AUROC $\uparrow$ & AURAC $\uparrow$ & Bal. Acc $\uparrow$ & N-tok $\downarrow$
  & AUROC $\uparrow$ & AURAC $\uparrow$ & Bal. Acc $\uparrow$ & N-tok $\downarrow$
  & AUROC $\uparrow$ & AURAC $\uparrow$ & Bal. Acc $\uparrow$ & N-tok $\downarrow$ \\
\midrule

\multirow{8}{*}{\rotatebox[origin=c]{90}{Gemma4-4B}}
& Semantic Entropy
    & 0.676$_{.006}$ & 0.443$_{.007}$ & 0.638$_{.006}$ & 173.0
    & 0.612$_{.009}$ & 0.583$_{.009}$ & 0.585$_{.007}$ & 209.7
    & 0.671$_{.003}$ & 0.708$_{.003}$ &  0.627$_{.003}$ & 216.1 \\
& Self-Consistency
    & 0.667$_{.006}$ & 0.445$_{.007}$ & 0.620$_{.005}$ & 173.0
    & 0.612$_{.009}$ & 0.585$_{.009}$ & 0.577$_{.008}$  & 209.7
    & 0.655$_{.003}$ & 0.703$_{.003}$ &  0.615$_{.003}$ & 216.1 \\
\cmidrule(lr){2-14}
& Sequence Entropy
    & 0.649$_{.006}$ & 0.436$_{.007}$ & 0.605$_{.006}$ & 17.4
    & 0.462$_{.009}$ & 0.505$_{.001}$ & 0.511$_{.003}$ & 20.8
    & 0.745$_{.003}$ & 0.745$_{.003}$ & 0.689$_{.003}$ & 21.5 \\
& LogTokU
    & 0.723$_{.006}$ & 0.484$_{.007}$ &  \underline{0.681}$_{.005}$& 17.4
    & \textbf{0.711}$_{.008}$ & 0.650$_{.008}$ & 0.640$_{.007}$ & 20.8
    & 0.524$_{.004}$ & 0.607$_{.003}$ & 0.544$_{.002}$ & 21.5 \\
& Hidden-State Score
    & 0.654$_{.006}$ & 0.445$_{.007}$ & 0.665$_{.005}$ & 17.4
    & 0.696$_{.008}$ & \textbf{0.656}$_{.008}$ & 0.643$_{.008}$ & 20.8
    & 0.373$_{.004}$ & 0.515$_{.004}$ &  0.506$_{.001}$& 21.5 \\
& Self-Certainty
    & 0.660$_{.006}$ & 0.445$_{.007}$ & 0.661$_{.005}$ & 17.4
    &0.511$_{.010}$ &0.542$_{.009}$ &  0.535$_{.006}$& 20.8
    & \underline{0.768}$_{.003}$ & \underline{0.755}$_{.006}$ & \underline{0.708}$_{.003}$  & 21.5 \\
\cmidrule(lr){2-14}
& Logit Magnitude ($\tau=T$)
    & \textbf{0.733}$_{.006}$ & \textbf{0.488}$_{.007}$ & \textbf{0.682}$_{.005}$ & 17.4
    & \underline{0.704}$_{.008}$ & \textbf{0.656}$_{.008}$ & \textbf{0.655}$_{.007}$ & 20.8 
    & \textbf{0.775}$_{.003}$ & \textbf{0.764}$_{.002}$ & \textbf{0.711}$_{.003}$ & 21.5 \\
& Logit Magnitude 
    & \underline{0.730}$_{.006}$ & \underline{0.486}$_{.005}$ & 0.675$_{.005}$  & 15.0
    & 0.699$_{.008}$ & \underline{0.652}$_{.008}$ & \underline{0.648}$_{.007}$ & 16.0
    & 0.740$_{.003}$ & 0.750$_{.003}$ & 0.687$_{.003}$ & 17.8 \\
& MetaUE
    & 0.709$_{.004}$ & 0.475$_{.003}$ &  0.656$_{.003}$ & 0
    & 0.617$_{.003}$ & 0.599$_{.002}$ & 0.579$_{.004}$ & 0
    & 0.733$_{.004}$ & 0.588$_{.002}$ & 0.667$_{.003}$ & 0 \\

\midrule

\multirow{8}{*}{\rotatebox[origin=c]{90}{Qwen3.5-4B}}
& Semantic Entropy
    & 0.754$_{.005}$ & 0.575$_{.007}$ & 0.702$_{.005}$ & 195.9
    & \underline{0.761}$_{.009}$ & \underline{0.798}$_{.006}$ & \textbf{0.714}$_{.008}$ &  161.1
    & 0.667$_{.003}$ & 0.547$_{.003}$ &  0.618$_{.003}$ & 430.2 \\
& Self-Consistency
    & 0.714$_{.006}$ & 0.561$_{.007}$ & 0.651$_{.005}$ & 195.9
    & \textbf{0.766}$_{.009}$ & \textbf{0.807}$_{.006}$ & 0.702$_{.008}$ & 161.1
    & 0.601$_{.003}$ & 0.506$_{.003}$ & 0.567$_{.003}$ & 430.2 \\
\cmidrule(lr){2-14}
& Sequence Entropy
    & 0.680$_{.006}$ & 0.542$_{.007}$ & 0.631$_{.005}$ & 19.4
    & 0.648$_{.010}$ & 0.756$_{.007}$ & 0.605$_{.008}$ & 15.7
    & 0.765$_{.003}$ & 0.600$_{.003}$ & 0.698$_{.002}$ & 43.3 \\
& LogTokU
    & 0.771$_{.005}$ & 0.587$_{.006}$ & 0.694$_{.005}$ & 19.4
    & 0.742$_{.005}$ & 0.779$_{.006}$ & 0.696$_{.008}$ & 15.7
    & 0.718$_{.003}$ & 0.565$_{.003}$ & 0.670$_{.002}$ & 43.3 \\
& Hidden-State Score
    & 0.704$_{.006}$ & 0.543$_{.007}$ & 0.658$_{.005}$ & 19.4
    & 0.651$_{.010}$ & 0.745$_{.008}$ & 0.598$_{.009}$ & 15.7
    & 0.719$_{.003}$ & 0.569$_{.003}$ & 0.671$_{.003}$ & 43.3 \\
& Self-Certainty
    & 0.651$_{.006}$ & 0.52$_{.007}$ & 0.614$_{.005}$ & 19.4
    & 0.566$_{.010}$ & 0.713$_{.008}$ & 0.548$_{.009}$ & 15.7
    & \underline{0.771}$_{.003}$ & 0.603$_{.003}$ & 0.707$_{.003}$ & 43.3 \\
\cmidrule(lr){2-14}
& Logit Magnitude ($\tau=T$)
    & \textbf{0.793}$_{.005}$ & \textbf{0.604}$_{.006}$ & \textbf{0.728}$_{.005}$ & 19.4
    & 0.751$_{.009}$ & 0.794$_{.007}$ & \underline{0.705}$_{.008}$ & 15.7
    & \textbf{0.775}$_{.008}$ & \textbf{0.605}$_{.008}$ & \textbf{0.713}$_{.002}$ & 43.3 \\
& Logit Magnitude
    & \underline{0.791}$_{.005}$ & \underline{0.603}$_{.005}$ &  \underline{0.727}$_{.005}$ & 16.3
    & 0.750$_{.009}$ & 0.794$_{.007}$ & 0.699$_{.008}$ & 11.7
    & 0.770$_{.003}$ & \underline{0.604}$_{.003}$ & \underline{0.711}$_{.002}$ & 38.2 \\
& MetaUE
    & 0.724$_{.003}$ & 0.564$_{.001}$ & 0.673$_{.004}$ & 0
    &0.717$_{.002}$ & 0.788$_{.001}$ & 0.656$_{.008}$ & 0
    & 0.733$_{.004}$ & 0.588$_{.002}$ & 0.667$_{.003}$ & 0 \\
    
\bottomrule
\end{tabular}}
\end{table*}

\subsection{Dynamics of Logit Magnitude for Partial-Generation Estimation}

\begin{figure}[h]
    \centering

    \begin{subfigure}[t]{0.325\textwidth}
        \centering
        \includegraphics[width=\linewidth]{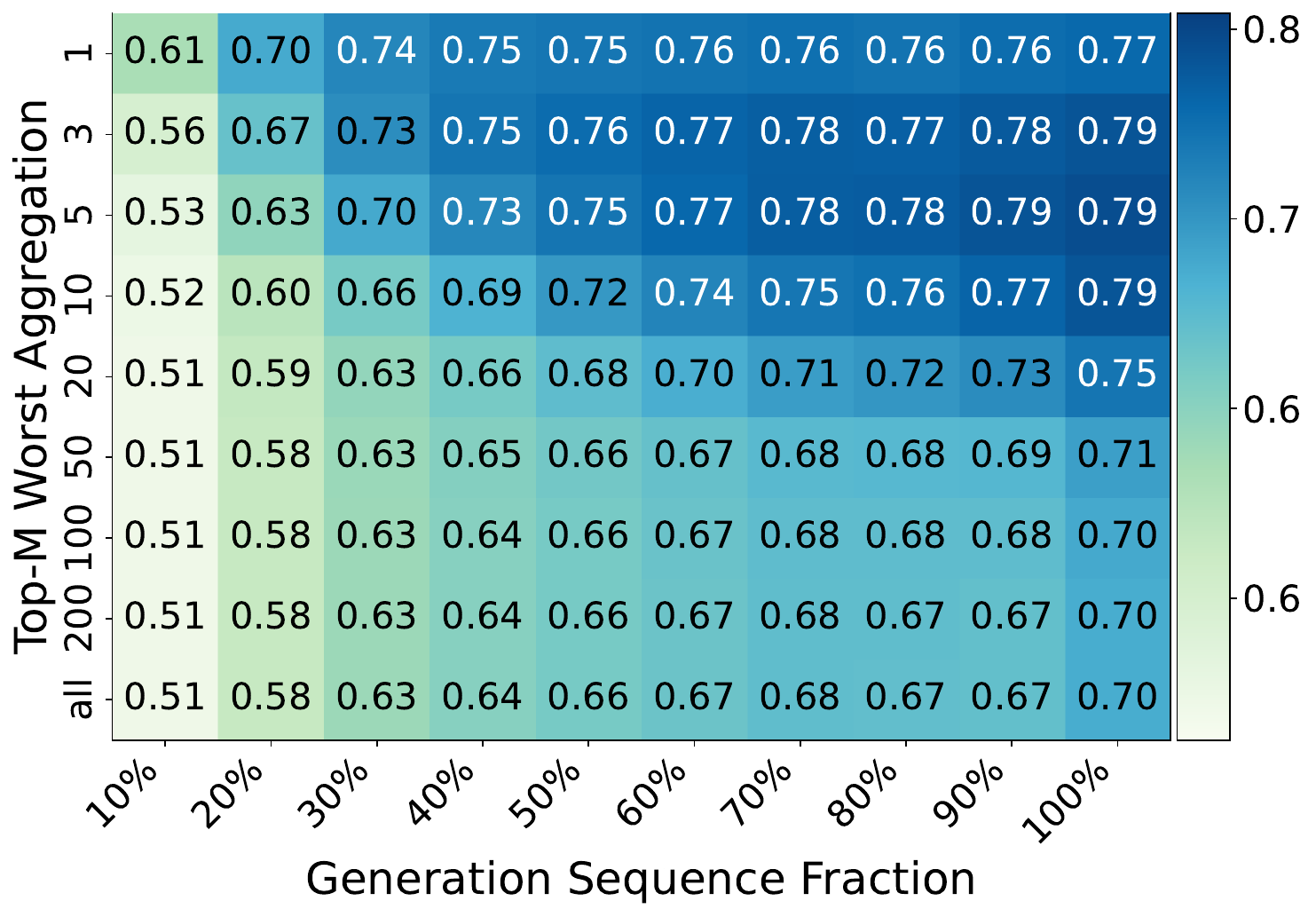}
        \caption{COQA, Qwen3.5}
    \end{subfigure}
    \hfill
    \begin{subfigure}[t]{0.325\textwidth}
        \centering
        \includegraphics[width=\linewidth]{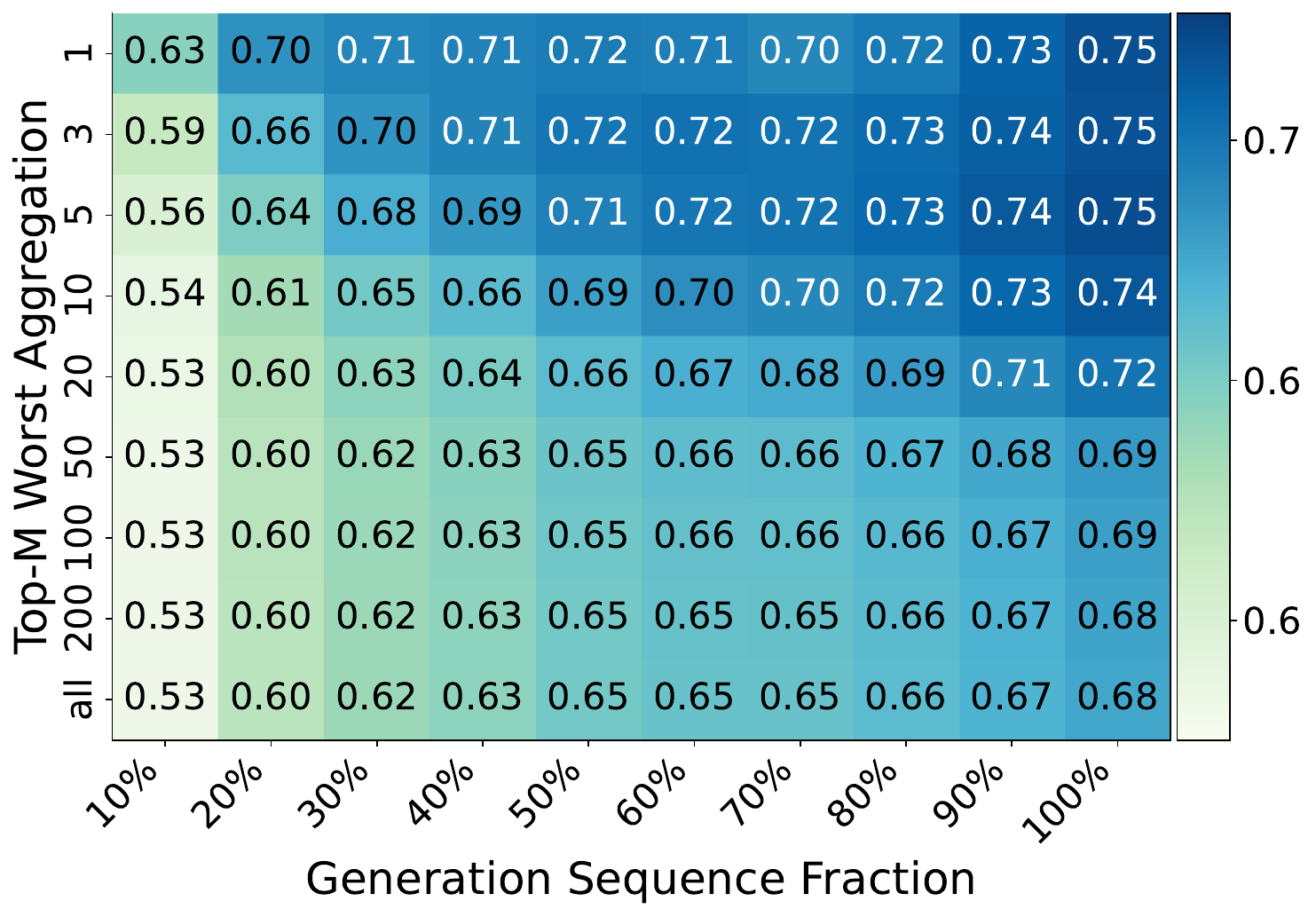}
        \caption{NewsQA, Qwen3.5}
    \end{subfigure}
    \hfill
    \begin{subfigure}[t]{0.325\textwidth}
        \centering
        \includegraphics[width=\linewidth]{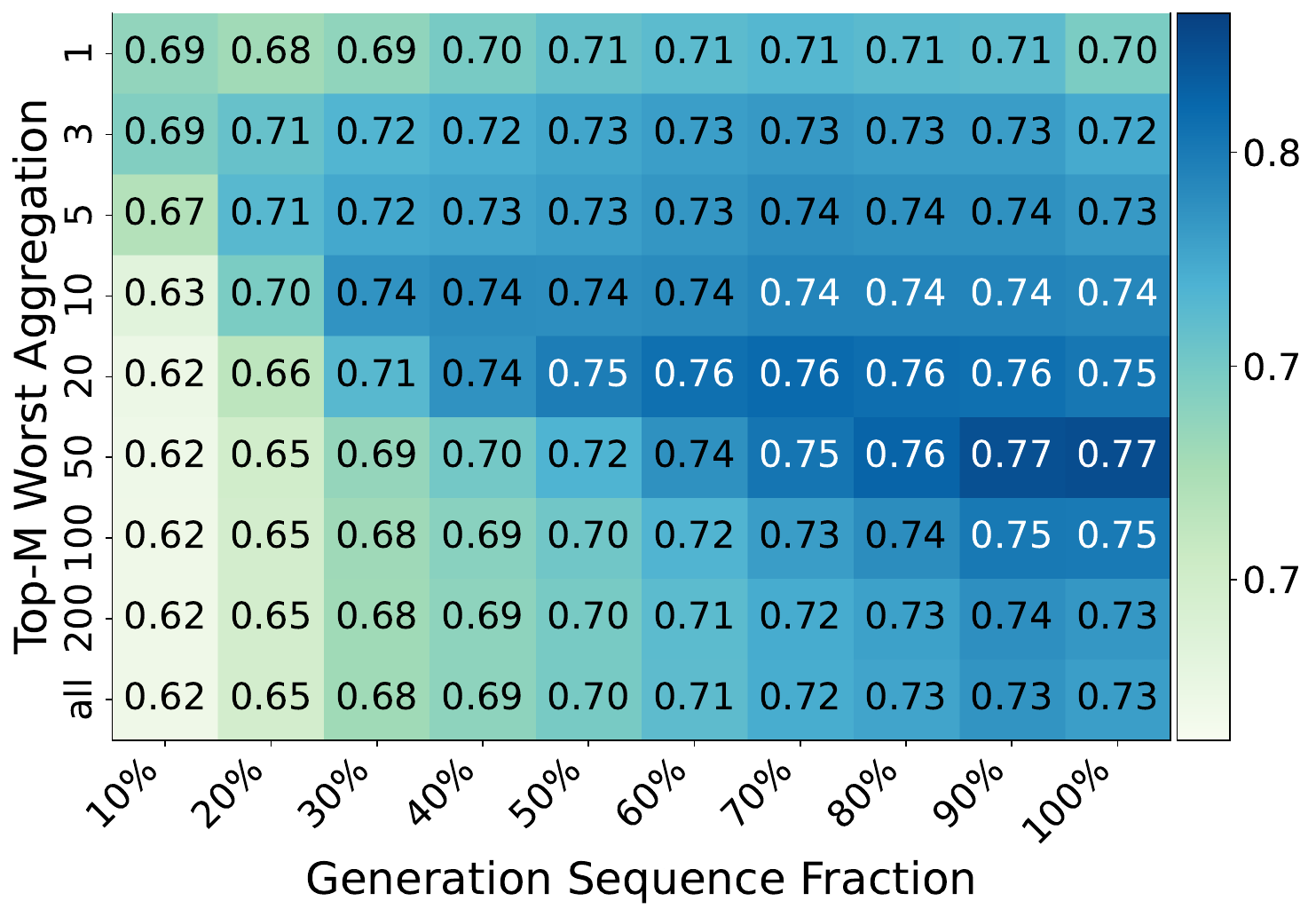}
        \caption{emrQA, Qwen3.5}
    \end{subfigure}

    \caption{
    Partial-generation dynamics of Logit Magnitude uncertainty estimation. AUROC is reported under different observed generation fractions and top-$M$ worst-token aggregation choices.}
\label{fig:partial_generation_dynamics}
\end{figure}

We next examine how the reliability of Logit Magnitude changes as more of the generation is observed. For each generated answer, we recompute the sequence-level uncertainty score $U_\tau$ using only the token-level scores $u_{1:\tau}$ available up to stopping time $\tau$. To analyse the effect of partial observation, we set $\tau$ to fixed fractions of the full generation length $T$, ranging from $10\%$ to $100\%$, and vary the top-$M$ aggregation parameter. This experiment studies how the observed generation fraction and the number of selected high-uncertainty tokens jointly affect uncertainty estimation performance.

Figure~\ref{fig:partial_generation_dynamics} shows that reliable uncertainty estimation can often be achieved before observing the full response, but the required stopping time and aggregation size vary across datasets. On COQA and NewsQA, strong AUROC is obtained from relatively short prefixes, especially when using sparse top-$M$ aggregation. COQA performs well with small values such as $M=3$ to $10$, while NewsQA favours even smaller values such as $M=1$ to $5$. In contrast, emrQA benefits from larger aggregation windows, especially around $M=20$ to $50$, and its performance improves more gradually as the observed generation fraction increases. This suggests that medical-domain uncertainty is distributed across a broader set of specialised tokens, whereas general-domain uncertainty is often concentrated in a smaller subset of high-impact tokens. Overall, the results empirically support the use of early-stopped uncertainty estimation, and also show that the optimal stopping behaviour and top-$M$ choice are domain-dependent. Appendix~\ref{sec:mw_logit_magnitude} provides a sensitivity analysis of the hyperparameters ($M$ and $W$) of Logit Magnitude, showing that $M\in\{5,10\}$ gives a good default when no validation data is available, and $W\approx 2M$ provides a practical balance between estimation performance and cost.

\subsection{Sensitivity of MetaUE Encoder and Training Label Selection}

\begin{table*}[h]
\vspace{-0.25cm}
\centering
\caption{MetaUE encoder comparison with Logit Magnitude as training label.}
\footnotesize
\setlength{\tabcolsep}{2.0pt}
\renewcommand{\arraystretch}{1.08}
\label{table:metaue_encoder}
\resizebox{\textwidth}{!}{
\begin{tabular}{ll|ccc|ccc|ccc}
\toprule
\multirow{2}{*}{\textbf{LLM}}
& \multirow{2}{*}{\textbf{Encoder}}
& \multicolumn{3}{c|}{\textbf{CoQA}} & \multicolumn{3}{c|}{\textbf{NewsQA}} & \multicolumn{3}{c}{\textbf{emrQA}} \\
\cmidrule(lr){3-5} \cmidrule(lr){6-8} \cmidrule(lr){9-11}
 &  & AUROC $\uparrow$ & AURAC $\uparrow$ & Bal. Acc $\uparrow$ & AUROC $\uparrow$ & AURAC $\uparrow$ & Bal. Acc $\uparrow$ & AUROC $\uparrow$ & AURAC $\uparrow$ & Bal. Acc $\uparrow$ \\
\midrule

\multirow{6}{*}{\rotatebox[origin=c]{90}{Gemma4-4B}}
& BGE-M3
    & 0.634$_{.004}$ & 0.424$_{.002}$ & 0.597$_{.010}$ & 0.561$_{.003}$ & 0.566$_{.002}$ & 0.541$_{.009}$ & 0.700$_{.003}$ & 0.732$_{.002}$ & 0.647$_{.003}$ \\
\cmidrule{2-11}
& Qwen3-Emb-0.6B
    & 0.660$_{.003}$ & 0.444$_{.002}$ & 0.617$_{.004}$ & 0.574$_{.003}$ & 0.574$_{.002}$ & 0.548$_{.004}$ & 0.714$_{.001}$ & 0.739$_{.001}$ & 0.658$_{.001}$ \\
    
& Qwen3-Emb-4B
    & 0.629$_{.003}$ & 0.423$_{.002}$ & 0.596$_{.002}$ & 0.562$_{.002}$ & 0.563$_{.002}$ & 0.543$_{.011}$ & 0.710$_{.005}$ & 0.738$_{.003}$ & 0.654$_{.004}$ \\

& Qwen3-Emb-8B
    & 0.689$_{.004}$ & 0.461$_{.002}$ & 0.640$_{.003}$ & 0.605$_{.007}$ & 0.589$_{.004}$ & 0.571$_{.004}$ & 0.716$_{.002}$ & 0.741$_{.001}$ & 0.659$_{.002}$ \\
\cmidrule{2-11}
& Qwen3-VL-2B
    & 0.709$_{.004}$ & 0.475$_{.003}$ & 0.656$_{.003}$ & 0.617$_{.003}$ & 0.599$_{.002}$ & 0.579$_{.005}$ & 0.719$_{.003}$ & 0.742$_{.002}$ & 0.663$_{.002}$ \\
& Qwen3-VL-8B
    & 0.726$_{.003}$ & 0.483$_{.002}$ & 0.668$_{.003}$ & 0.636$_{.003}$ & 0.613$_{.002}$ & 0.606$_{.004}$ & 0.716$_{.003}$ & 0.741$_{.001}$ & 0.661$_{.003}$ \\
\cmidrule(lr){1-11}

\multirow{6}{*}{\rotatebox[origin=c]{90}{Qwen3.5-4B}}
& BGE-M3
    & 0.673$_{.002}$ & 0.530$_{.001}$ & 0.630$_{.002}$ & 0.632$_{.006}$ & 0.749$_{.002}$ & 0.595$_{.006}$ & 0.702$_{.005}$ & 0.570$_{.003}$ & 0.642$_{.005}$ \\
\cmidrule{2-11}
& Qwen3-Emb-0.6B
    & 0.688$_{.002}$ & 0.542$_{.000}$ & 0.641$_{.002}$ & 0.650$_{.006}$ & 0.757$_{.003}$ & 0.612$_{.005}$ & 0.726$_{.002}$ & 0.585$_{.001}$ & 0.659$_{.002}$ \\

& Qwen3-Emb-4B
    & 0.661$_{.002}$ & 0.525$_{.001}$ & 0.616$_{.003}$ & 0.618$_{.002}$ & 0.740$_{.001}$ & 0.581$_{.003}$ & 0.722$_{.007}$ & 0.583$_{.004}$ & 0.657$_{.005}$ \\

& Qwen3-Emb-8B
    & 0.716$_{.001}$ & 0.560$_{.001}$ & 0.661$_{.003}$ & 0.688$_{.004}$ & 0.774$_{.002}$ & 0.638$_{.004}$ & 0.730$_{.006}$ & 0.588$_{.004}$ & 0.667$_{.005}$ \\
\cmidrule{2-11}
& Qwen3-VL-2B
    & 0.724$_{.003}$ & 0.564$_{.001}$ & 0.673$_{.004}$ & 0.717$_{.002}$ & 0.788$_{.001}$ & 0.656$_{.008}$ & 0.733$_{.004}$ & 0.588$_{.002}$ & 0.667$_{.003}$ \\

& Qwen3-VL-8B
    & 0.733$_{.003}$ & 0.568$_{.002}$ & 0.680$_{.003}$ & 0.736$_{.003}$ & 0.797$_{.001}$ & 0.679$_{.004}$ & 0.733$_{.001}$ & 0.590$_{.001}$ & 0.668$_{.002}$ \\

\bottomrule
\end{tabular}}
\end{table*}

\begin{figure*}[ht]
    \centering
    \begin{subfigure}[t]{0.325\textwidth}
        \centering
        \includegraphics[width=\linewidth]{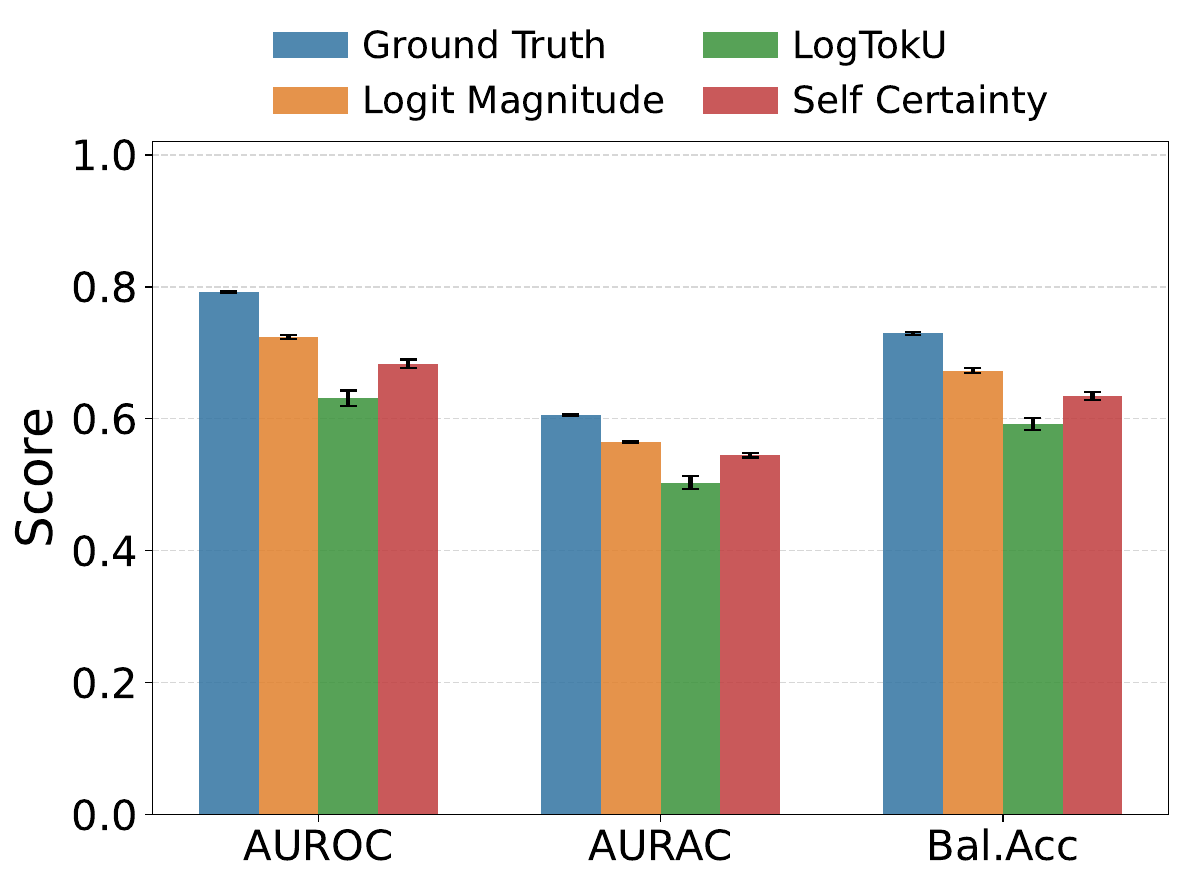}
        \caption{COQA, Qwen3.5}
    \end{subfigure}
    \hfill
    \begin{subfigure}[t]{0.325\textwidth}
        \centering
        \includegraphics[width=\linewidth]{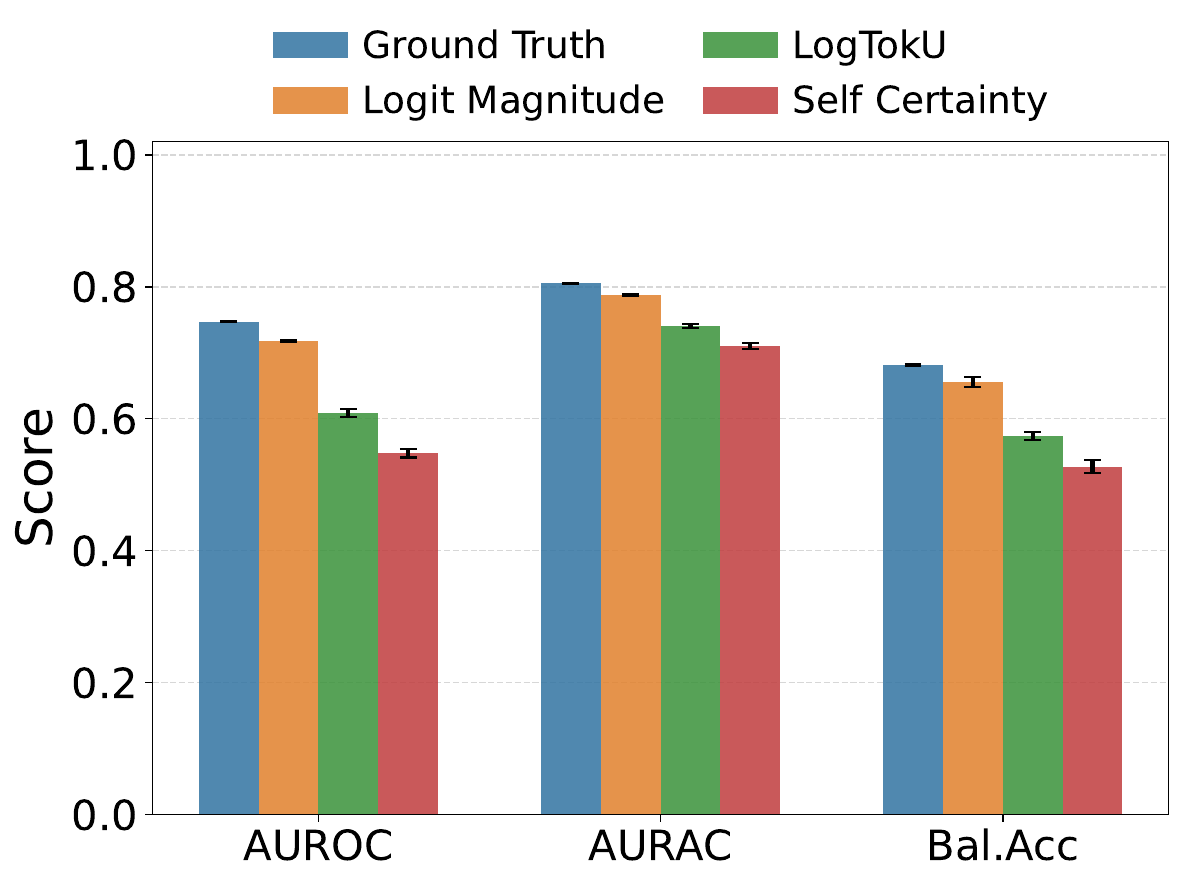}
        \caption{NewsQA, Qwen3.5}
    \end{subfigure}
    \hfill
    \begin{subfigure}[t]{0.325\textwidth}
        \centering
        \includegraphics[width=\linewidth]{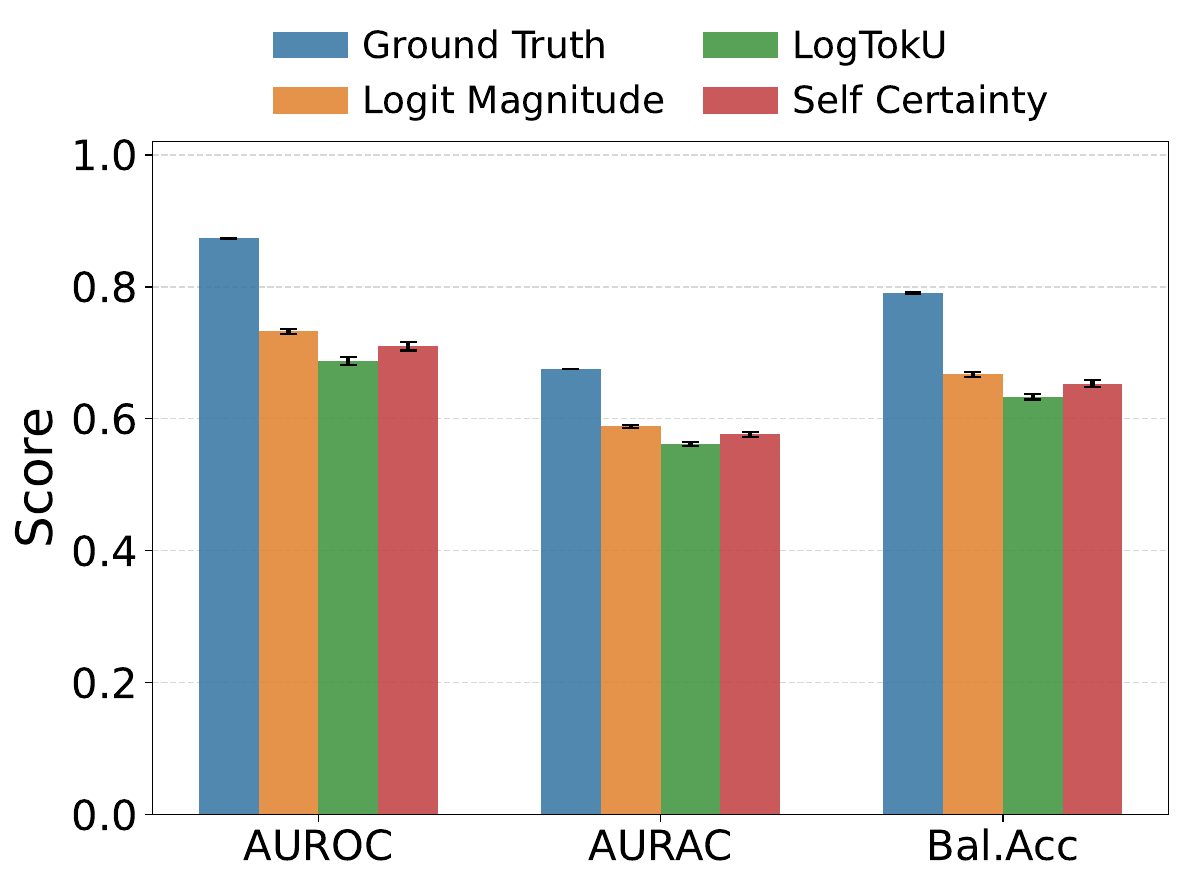}
        \caption{emrQA, Qwen3.5}
    \end{subfigure}
    \caption{
    MetaUE design analysis on Qwen3.5-4B. Figure (a-c) compares different training signals with Qwen3-VL-Embedding-2B fixed as the encoder. 
    }
    \label{fig:MetaUE_design}
    \vspace{-0.25cm}
\end{figure*}

This experiment investigates two design factors for MetaUE. First, we fix the training label to Logit Magnitude and vary the frozen encoder backbone, including BGE-M3~\cite{chen2024m3}, Qwen3-Embedding~\cite{zhang2025qwen3}, and Qwen3-VL-Embedding~\cite{li2026qwen3} at different model scales. Second, we fix the encoder to Qwen3-VL-Embedding-2B and compare four training signals, including ground-truth correctness, Logit Magnitude, LogTokU, and self-certainty. As shown in Table~\ref{table:metaue_encoder}, Qwen3-VL-Embedding provides consistently strong performance across both LLMs and all three datasets. Although Qwen3-VL-Embedding-8B often gives the highest scores in most settings, the 2B model already achieves competitive performance, making it a practical trade-off between estimation quality and computational cost. We therefore use Qwen3-VL-Embedding-2B as the default encoder for MetaUE.
As shown in Figure~\ref{fig:MetaUE_design}(a-c), the ground-truth label gives the strongest reference performance, since it directly uses dataset correctness supervision. Among uncertainty-derived pseudo-labels, Logit Magnitude achieves the best overall performance across all datasets, consistently outperforming LogTokU and self-certainty. This suggests that generation-level logit evidence provides a more effective supervision target for input-only uncertainty prediction than probability-based or self-evaluation signals.
Beyond design analyses, Appendix~\ref{append_sec:transfer} evaluates MetaUE's transferability across datasets and LLM architectures, providing evidence on the robustness of the MetaUE when transferring.

\subsection{Generalisation of Logit Magnitude and MetaUE}
\begin{table*}[h]
\vspace{-0.25cm}
\centering
\caption{
Assessment of uncertainty estimation methods across LLM families and sizes. 
The $^*$ symbol indicates a Mixture-of-Experts model. \textbf{Bold} values indicate the best performance.
}
\footnotesize
\setlength{\tabcolsep}{2.0pt}
\renewcommand{\arraystretch}{1.08}
\label{table:scalability_comparison}
\resizebox{\textwidth}{!}{
\begin{tabular}{lll|cccc|cccc|cccc}
\toprule
\multirow{2}{*}{\textbf{LLM}}
& \multirow{2}{*}{\textbf{Size}}
& \multirow{2}{*}{\textbf{Method}}
& \multicolumn{4}{c|}{\textbf{COQA}}
& \multicolumn{4}{c|}{\textbf{NewsQA}}
& \multicolumn{4}{c}{\textbf{emrQA}} \\
\cmidrule(lr){4-7} \cmidrule(lr){8-11} \cmidrule(lr){12-15}
& & & AUROC $\uparrow$ & AURAC $\uparrow$ & Bal. Acc $\uparrow$ & N-tok $\downarrow$
  & AUROC $\uparrow$ & AURAC $\uparrow$ & Bal. Acc $\uparrow$ & N-tok $\downarrow$
  & AUROC $\uparrow$ & AURAC $\uparrow$ & Bal. Acc $\uparrow$ & N-tok $\downarrow$ \\
\midrule

\multirow{8}{*}{\rotatebox[origin=c]{90}{Gemma4}}
& \multirow{4}{*}{26B$^*$}
& LogTokU
    & 0.822$_{.005}$ & 0.695$_{.006}$ & 0.736$_{.005}$ & 13.8
    & 0.719$_{.010}$ & 0.817$_{.006}$ & 0.657$_{.009}$ & 9.7
    & 0.672$_{.003}$ & 0.671$_{.003}$ & 0.658$_{.003}$ & 21.5 \\
& & Self-Certainty
    & 0.675$_{.006}$ & 0.599$_{.007}$ & 0.625$_{.005}$ & 13.8
    & 0.632$_{.011}$ & 0.782$_{.007}$ & 0.592$_{.008}$ & 9.7
    & 0.729$_{.003}$ & 0.714$_{.003}$ & \textbf{0.687}$_{.003}$ & 21.5 \\
\cmidrule{3-15}
& & Logit Magnitude
    & \textbf{0.830}$_{.005}$ & \textbf{0.699}$_{.006}$ & \textbf{0.756}$_{.005}$ & 13.8
    & \textbf{0.755}$_{.009}$ & \textbf{0.833}$_{.005}$ & \textbf{0.692}$_{.008}$ & 9.7
    & \textbf{0.731}$_{.003}$ & \textbf{0.722}$_{.003}$ & 0.675$_{.003}$ & 21.5 \\
& & MetaUE
    & 0.794$_{.003}$ & 0.676$_{.001}$ & 0.722$_{.005}$ & 0
    & 0.714$_{.004}$ & 0.822$_{.002}$ & 0.648$_{.007}$ & 0
    & 0.696$_{.007}$ & 0.707$_{.004}$ & 0.648$_{.005}$ & 0 \\
\cmidrule(lr){2-15}
& \multirow{4}{*}{31B}
& LogTokU
    & 0.819$_{.005}$ & 0.704$_{.005}$ & 0.735$_{.005}$ & 13.5
    & 0.694$_{.011}$ & 0.823$_{.006}$ & 0.647$_{.010}$ & 8.9
    & 0.512$_{.003}$ & 0.514$_{.003}$ & 0.572$_{.002}$ & 19.5 \\
& & Self-Certainty
    & 0.710$_{.006}$ & 0.627$_{.006}$ & 0.649$_{.005}$ & 13.5
    & 0.663$_{.011}$ & 0.813$_{.006}$ & 0.608$_{.010}$ & 8.9
    & 0.753$_{.003}$ & 0.705$_{.003}$ & \textbf{0.714}$_{.002}$ & 19.5 \\
    \cmidrule{3-15}
& & Logit Magnitude
    & \textbf{0.855}$_{.004}$ & \textbf{0.721}$_{.005}$ & \textbf{0.781}$_{.004}$ & 13.5
    & \textbf{0.759}$_{.009}$ & \textbf{0.846}$_{.005}$ & \textbf{0.697}$_{.008}$ & 8.9
    & \textbf{0.761}$_{.003}$ & \textbf{0.710}$_{.003}$ & 0.703$_{.002}$ & 19.5 \\
& & MetaUE
    & 0.807$_{.004}$ & 0.692$_{.002}$ & 0.739$_{.005}$ & 0
    & 0.717$_{.002}$ & 0.836$_{.001}$ & 0.659$_{.004}$ & 0
    & 0.717$_{.001}$ & 0.685$_{.001}$ & 0.662$_{.001}$ & 0 \\
\midrule

\multirow{8}{*}{\rotatebox[origin=c]{90}{Qwen3.5}}
& \multirow{4}{*}{9B}
& LogTokU
    & 0.781$_{.005}$ & 0.652$_{.006}$ & 0.704$_{.005}$ & 18.6
    & 0.749$_{.008}$ & 0.744$_{.006}$ & \textbf{0.689}$_{.008}$ & 24.1
    & 0.696$_{.003}$ & 0.612$_{.003}$ & 0.664$_{.003}$ & 41.1 \\
& & Self-Certainty
    & 0.608$_{.006}$ & 0.544$_{.007}$ & 0.578$_{.005}$ & 18.6
    & 0.548$_{.010}$ & 0.669$_{.008}$ & 0.534$_{.008}$ & 24.1
    & 0.729$_{.003}$ & 0.638$_{.003}$ & 0.680$_{.003}$ & 41.1 \\
    \cmidrule{3-15}
& & Logit Magnitude
    & \textbf{0.791}$_{.005}$ & \textbf{0.656}$_{.006}$ & 0.\textbf{723}$_{.005}$ & 18.6
    & \textbf{0.753}$_{.009}$ & \textbf{0.772}$_{.007}$ & 0.686$_{.008}$ & 24.1
    & \textbf{0.733}$_{.003}$ & \textbf{0.641}$_{.003}$ & \textbf{0.683}$_{.004}$ & 41.1 \\
& & MetaUE
    & 0.701$_{.004}$ & 0.601$_{.002}$ & 0.652$_{.004}$ & 0
    & 0.679$_{.002}$ & 0.742$_{.002}$ & 0.630$_{.002}$ & 0
    & 0.710$_{.003}$ & 0.642$_{.002}$ & 0.648$_{.002}$ & 0 \\
\cmidrule(lr){2-15}
& \multirow{4}{*}{35B$^*$}
& LogTokU
    & 0.816$_{.005}$ & 0.663$_{.006}$ & 0.735$_{.005}$ & 24.4
    & \textbf{0.727}$_{.008}$ & 0.751$_{.006}$ & 0.667$_{.008}$ & 16.3
    & 0.583$_{.003}$ & 0.430$_{.004}$ & 0.613$_{.002}$ & 38.0 \\
& & Self-Certainty
    & 0.591$_{.006}$ & 0.528$_{.007}$ & 0.563$_{.005}$ & 24.4
    & 0.553$_{.010}$ & 0.692$_{.008}$ & 0.534$_{.007}$ & 16.3
    & 0.715$_{.003}$ & \textbf{0.558}$_{.003}$ & 0.682$_{.002}$ & 38.0 \\
\cmidrule{3-15}
& & Logit Magnitude
    & \textbf{0.857}$_{.004}$ & \textbf{0.685}$_{.006}$ & \textbf{0.785}$_{.004}$ & 24.4
    & 0.722$_{.009}$ & \textbf{0.774}$_{.007}$ & \textbf{0.671}$_{.008}$ & 16.3
    & \textbf{0.720}$_{.003}$ & 0.554$_{.003}$ & \textbf{0.687}$_{.003}$ & 38.0 \\
& & MetaUE
    & 0.765$_{.002}$ & 0.638$_{.001}$ & 0.701$_{.003}$ & 0
    & 0.686$_{.001}$ & 0.763$_{.001}$ & 0.637$_{.003}$ & 0
    & 0.702$_{.003}$ & 0.551$_{.001}$ & 0.646$_{.002}$ & 0 \\
\bottomrule
\end{tabular}}
\end{table*}

To evaluate whether the proposed methods generalise beyond the 4B models, we extend the comparison to larger LLMs from Gemma4 and Qwen3.5. We also include Mixture-of-Experts (MoE) models~\cite{cai2025survey} to test whether the proposed methods remain effective under different architectural designs. 
As shown in Table~\ref{table:scalability_comparison}, Logit Magnitude achieves the best or near-best AUROC and AURAC across most Gemma4 and Qwen3.5 settings. The gains are consistent across both general-domain and medical-domain datasets, indicating that positive logit evidence remains a strong uncertainty signal for larger dense and MoE models. MetaUE also generalises well across these model families, often approaching generation-based baselines while requiring zero generated tokens. On Qwen3.5, MetaUE is particularly competitive on emrQA, where it achieves AURAC $0.642$ for the 9B model and $0.551$ for the 35B MoE model. These results suggest that both proposed methods scale beyond the main 4B setting. Additional results on Llama3 are provided in Appendix~\ref{app_sec:evaluation on llama3}.

\section{Conclusion}
\label{sec:conclusion}
In this paper, we investigated how much generation is needed for effective LLM uncertainty estimation. By formulating uncertainty estimation as an early estimation problem over the generation process, we proposed Logit Magnitude with top-$M$ aggregation and an early-stop mechanism and MetaUE for input-only uncertainty estimation. Across general-domain and medical-domain benchmarks, as well as larger dense and MoE models, Logit Magnitude achieves strong performance, and MetaUE provides a useful zero-generation approximation. These results suggest that uncertainty estimation should focus on compact, informative signals of unreliability rather than exhaustive generation.

This work has several limitations. First, our evaluation focuses on open-answer question generation, following the common evaluation protocol in LLM uncertainty estimation. Other task formats, such as multiple-choice question answering, multi-selection question answering, and long-form generation, remain underexplored. Second, we observe a failure mode of Logit Magnitude on Llama3 in the emrQA setting and attribute it to poorly calibrated logit distributions. Although this issue does not appear on the latest LLMs with well-designed calibration during post-training, the current study does not provide a comprehensive analysis of when this failure mode occurs. Future work should extend evaluation to broader generation tasks and systematically study the relationship between logit calibration, domain shift, and uncertainty estimation performance.

\bibliography{reference}
\bibliographystyle{unsrt}

\newpage
\appendix

\section{Algorithm for Logit Magnitude with Adaptive Early Stopping}
\label{app:algorithm_logit_magnitude}

Algorithm~\ref{alg:logit_magnitude_early_stop} summarises the proposed Logit Magnitude estimator with top-$M$ aggregation and adaptive early stopping.

\begin{algorithm}[h!]
\caption{Logit Magnitude with Top-$M$ Aggregation and Early Stopping}
\label{alg:logit_magnitude_early_stop}
\begin{algorithmic}[1]
\Require LLM $\theta$, input prompt $\mathbf{z}$, top-logit size $K$, aggregation size $M$, patience window $W$.
\Ensure Logit Magnitude score $\widetilde{U}_{\tau}$.

\State Initialise generated prefix $\mathbf{y}_{1:0}\gets \emptyset$, top-$M$ score set $\mathcal{H}\gets \emptyset$, patience counter $n\gets 0$, and time step $t\gets 0$.

\While{\textbf{true}}
    \State $t\gets t+1$
    \State Obtain logits $\mathbf{\ell}_t$ from $\theta(\mathbf{z},\mathbf{y}_{<t})$ 
    \State Generate next token $y_t$ using the decoding strategy
    \State Append $y_t$ to the generated prefix $\mathbf{y}_{1:t}$
    \State Compute the token-level Logit Magnitude score with Eq~\ref{eq:token_level_logit_magnitude}.

    \If{$|\mathcal{H}| < M$}
        \State Add $(t,u_t)$ to $\mathcal{H}$
        \State $n\gets 0$
    \ElsIf{$u_t > \min_{(j,u_j)\in\mathcal{H}} u_j$}
        \State Remove $\arg\min_{(j,u_j)\in\mathcal{H}} u_j$ from $\mathcal{H}$
        \State Add $(t,u_t)$ to $\mathcal{H}$
        \State $n\gets 0$
    \Else
        \State $n\gets n+1$
    \EndIf

    \If{$y_t$ is an end-of-sequence token \textbf{or} $(|\mathcal{H}|=M \text{ and } n\geq W)$}
    \State \textbf{break}
    \EndIf
\EndWhile

\State Compute the sequence-level uncertainty score $\widetilde{U}_{\tau}$ with Eq~\ref{eq:sequence_level_logit_magnitude}. 
\State Apply min-max normalisation using training-set statistics.

\State \Return $\widetilde{U}_{\tau}$
\end{algorithmic}
\end{algorithm}

\begin{algorithm}[h!]
\caption{MetaUE Training}
\label{alg:metaue_training}
\begin{algorithmic}[1]
\Require Training prompts $\mathcal{D}_{\mathrm{train}}=\{\mathbf{z}_i\}_{i=1}^{n}$, frozen LLM $\theta$, frozen encoder $\phi$, trainable MLP head $g_{\omega}$, learning rate $\alpha$, Logit Magnitude scores $\mathbf{U}_{\tau_i}$ collected using Algorithm~\ref{alg:logit_magnitude_early_stop}.
\Ensure Trained MetaUE model $f_{\omega}(\mathbf{z})=g_{\omega}(\phi(\mathbf{z}))$

\For{$e=1,\ldots,E$}
    \For{each mini-batch $\mathcal{B}\subseteq\mathcal{S}$ with size $B$}
        \State Encode prompts with the frozen encoder:
        \[
            \mathbf{h}_i=\phi(\mathbf{z}_i), \qquad \mathbf{z}_i\in\mathcal{B}.
        \]
        \State Predict input-only uncertainty scores with Eq~\ref{eq:MetaUE}.
        \State Compute the mean squared error loss $\mathcal{J}(\omega)$ with $\mathbf{U}_{\tau_i}$ using Eq~\ref{eq:MetaUE_mse_loss}.
        \State Update only the MLP parameters $\omega$ \[
            \omega
            \leftarrow
            \omega - \alpha \nabla_{\omega}\mathcal{J}(\omega),
        \]
    \EndFor
\EndFor

\State \Return trained MetaUE predictor $f_{\omega}$
\end{algorithmic}
\end{algorithm}

\section{Algorithm for MetaUE Training}
\label{app:algorithm_metaue}

Algorithm~\ref{alg:metaue_training} summarises the training procedure for MetaUE. The key idea is to first compute generation-based uncertainty scores using the Logit Magnitude, and then use the scores as pseudo-labels to train an input-only estimator.

\section{Derivation for the Error of Partial-Generation Uncertainty Estimation}
\label{app:early_stop_martingale}

Let $X\in L^2$ denote the scalar target to be estimated, such as final unreliability. Let $(\mathcal{F}_t)_{t=0}^{T}$ be the filtration generated by the prompt and the observed generation trajectory up to step $t$. The best estimate of $X$ after observing $\mathcal{F}_t$ is the conditional expectation
\begin{equation}
M_t=\mathbb{E}[X\mid\mathcal{F}_t].
\end{equation}

By the tower property of conditional expectation, $(M_t)_{t=0}^{T}$ is a Doob martingale~\cite{williams1991probability}. Define the martingale increment
\begin{equation}
\Delta_t=M_t-M_{t-1}.
\end{equation}
Then $\mathbb{E}[\Delta_t\mid\mathcal{F}_{t-1}]=0$.

Let $\tau\leq T$ be a stopping time. In our setting, $\tau$ corresponds to the time at which generation is stopped for uncertainty estimation. The penalty of stopping at $\tau$, relative to observing the full trajectory up to $T$, is
\begin{equation}
\mathcal{E}(\tau)
=
\mathbb{E}\!\left[(M_T-M_{\tau})^2\right].
\end{equation}
Since $T$ is finite, the difference between the full-sequence estimate and the early-stopped estimate can be written as the sum of the increments that are not observed:
\begin{equation}
M_T-M_{\tau}
=
\sum_{t=1}^{T}\Delta_t\mathbf{1}_{\{t>\tau\}} .
\end{equation}
Squaring this expression gives
\begin{equation}
(M_T-M_{\tau})^2
=
\sum_{t=1}^{T}\Delta_t^2\mathbf{1}_{\{t>\tau\}}
+
2\sum_{1\leq s<t\leq T}
\Delta_s\Delta_t
\mathbf{1}_{\{s>\tau\}}
\mathbf{1}_{\{t>\tau\}} .
\end{equation}

For $s<t$, the term
$\Delta_s\mathbf{1}_{\{s>\tau\}}\mathbf{1}_{\{t>\tau\}}$ is $\mathcal{F}_{t-1}$-measurable. First, because $s < t$ (and thus $s \leq t-1$), the prior increment $\Delta_s$ is already observed by time $t-1$ and is therefore $\mathcal{F}_{t-1}$-measurable. Second, the event $\{t > \tau\}$ is logically equivalent to $\{\tau \leq t-1\}$. By the strict definition of a stopping time, the event that stopping occurs at or before step $t-1$ is entirely determined by the filtration $\mathcal{F}_{t-1}$. Consequently, the indicator function $\mathbf{1}_{\{t>\tau\}}$ is $\mathcal{F}_{t-1}$-measurable. Similarly, since $s-1 < t-1$, the indicator $\mathbf{1}_{\{s>\tau\}}$ is also $\mathcal{F}_{t-1}$-measurable.

Since the entire product $\Delta_s\mathbf{1}_{\{s>\tau\}}\mathbf{1}_{\{t>\tau\}}$ is $\mathcal{F}_{t-1}$-measurable, it can be factored out of the inner conditional expectation. Applying the tower property of conditional expectation, we have:
\begin{align}
    \mathbb{E}\!\left[\Delta_s\Delta_t\mathbf{1}_{\{s>\tau\}}\mathbf{1}_{\{t>\tau\}}\right] 
    &= \mathbb{E}\!\left[\mathbb{E}\!\left[\Delta_s\Delta_t\mathbf{1}_{\{s>\tau\}}\mathbf{1}_{\{t>\tau\}} \mid \mathcal{F}_{t-1}\right]\right] \nonumber \\
    &= \mathbb{E}\!\left[\Delta_s\mathbf{1}_{\{s>\tau\}}\mathbf{1}_{\{t>\tau\}}\mathbb{E}[\Delta_t \mid \mathcal{F}_{t-1}]\right].
\end{align}

By the martingale property, the expected forward increment given the current history is exactly zero, i.e., $\mathbb{E}[\Delta_t \mid \mathcal{F}_{t-1}] = 0$. Substituting this into the expression naturally eliminates the entire term:
\begin{equation}
    \mathbb{E}\!\left[\Delta_s\mathbf{1}_{\{s>\tau\}}\mathbf{1}_{\{t>\tau\}} \cdot 0\right] = 0.
\end{equation}

Thus, only the squared increments remain:
\begin{equation}
\mathcal{E}(\tau)
=
\mathbb{E}\!\left[
\sum_{t=1}^{T}
\Delta_t^2\mathbf{1}_{\{t>\tau\}}
\right]
=
\mathbb{E}\!\left[
\sum_{t=\tau+1}^{T}
\Delta_t^2
\right].
\end{equation}

This identity shows that the early-stopping penalty is exactly the expected residual quadratic variation of the martingale after stopping. To obtain a simple bound, assume that the stopping rule is calibrated so that the unobserved tail has small conditional update energy:
\begin{equation}
\mathbb{E}[\Delta_t^2\mid\mathcal{F}_{t-1}]
\leq
\epsilon
\quad
\text{on the event } \{t>\tau\}.
\end{equation}
Since $\mathbf{1}_{\{t>\tau\}}$ is $\mathcal{F}_{t-1}$-measurable, we have
\begin{align}
\mathcal{E}(\tau)
&=
\mathbb{E}\!\left[
\sum_{t=1}^{T}
\mathbf{1}_{\{t>\tau\}}
\mathbb{E}[\Delta_t^2\mid\mathcal{F}_{t-1}]
\right] \nonumber \\
&\leq
\mathbb{E}\!\left[
\sum_{t=1}^{T}
\epsilon\,\mathbf{1}_{\{t>\tau\}}
\right] \nonumber \\
&=
\epsilon\,\mathbb{E}[T-\tau]
\leq
\epsilon T.
\end{align}

\section{Experiment Setup Details}
\label{app_sec:setup_details}

\subsection{Dataset Details}
\label{app:dataset_details}

In this section, we provide detailed descriptions of the three datasets utilised to evaluate our uncertainty estimation methods. The datasets were selected to cover a diverse range of domains (general news/conversational, and specialised medical data).

\textbf{COQA (Conversational Question Answering)~\cite{reddy2019coqa}:} 
A large-scale dataset for building conversational question-answering systems. It contains text passages from seven diverse domains alongside conversational questions and answers. COQA requires models to understand the pragmatics of conversation and correctly resolve coreferences across multiple turns of free-text generation.

\textbf{NewsQA~\cite{trischler2017newsqa}:} 
A challenging machine comprehension dataset containing over 100,000 human-generated question-answer pairs. The questions are based on CNN and Daily Mail news articles. It is designed to test a model's ability to perform complex reasoning, such as synthesising information from multiple sentences and handling lexical variations, making it an excellent benchmark for general-domain free-text uncertainty.

\textbf{emrQA~\cite{pampari2018emrqa}:} 
A large-scale medical question-answering dataset constructed from electronic medical records (EMRs). The input contexts are clinical notes that frequently contain fragmented, complex, or contradictory patient histories. The model is required to generate free-text answers. This dataset is uniquely suited for evaluating uncertainty estimation, as the underlying EMR data accurately reflects the noisy, ambiguous nature of real-world clinical decision-making.

\begin{table}[h]
\centering
\caption{Dataset statistics for train/validation/test splits.}
\label{tab:dataset_stats}
\begin{tabular}{llrrr}
\toprule
\textbf{Dataset} & \textbf{Domain} & \textbf{Train} & \textbf{Validation} & \textbf{Test} \\
\midrule
CoQA   & General    & 108,647 & 7,983 & 4,541 \\
NewsQA & General    & 92,549  & 5,166 & 5,126 \\
emrQA  & Healthcare & 320,142 & 40,017 & 40,018 \\
\bottomrule
\end{tabular}
\end{table}

\subsection{LLM Sampling Parameters}
\label{app:sampling_parameters}

To ensure a fair comparison across LLMs and uncertainty estimation methods, we use the same decoding configuration for all models and datasets. Specifically, generation is performed with the configuration of temperature $1.0$, top-$K=20$, and top-$p=1.0$. 

\subsection{Evaluation Metrics}
\label{app:evaluation_metrics}

In this section, we provide the formal definitions for the metrics used to evaluate uncertainty estimation methods. Let $\mathcal{D} = \{(z_i, y_i)\}_{i=1}^N$ be an evaluation dataset of size $N$, where each sample is associated with a continuous uncertainty score $u_i \in \mathbb{R}$ and a binary correctness label $c_i \in \{0,1\}$. Here, $c_i=1$ indicates that the generated answer is correct, while $c_i=0$ indicates that it is incorrect. We define the subset of correct generations as $\mathcal{D}_c = \{i \mid c_i = 1\}$ and the subset of incorrect generations as $\mathcal{D}_w = \{i \mid c_i = 0\}$.

\textbf{AUROC}
AUROC evaluates the threshold-independent discriminative power of an uncertainty estimator. Since higher uncertainty should correspond to lower correctness, AUROC represents the probability that a randomly selected incorrect generation is assigned a higher uncertainty score than a randomly selected correct generation:
\begin{equation}
    \mathrm{AUROC}
    =
    \frac{1}{|\mathcal{D}_c| |\mathcal{D}_w|}
    \sum_{i \in \mathcal{D}_c}
    \sum_{j \in \mathcal{D}_w}
    \mathbb{I}[u_j > u_i],
\end{equation}
where $\mathbb{I}[\cdot]$ is the indicator function. An AUROC of 0.5 corresponds to random ranking, while an AUROC of 1.0 corresponds to perfect separation between correct and incorrect responses.

\textbf{AURAC}
AURAC assesses the practical utility of uncertainty scores for selective prediction. It measures how the accuracy of the retained predictions changes as the most uncertain samples are progressively rejected. Let the dataset be sorted in ascending order of uncertainty, such that $u_{(1)} \leq u_{(2)} \leq \cdots \leq u_{(N)}$, with corresponding correctness labels $c_{(i)}$. If we reject the $k$ most uncertain samples, where $k \in \{0,1,\ldots,N-1\}$, the accuracy on the remaining accepted samples is
\begin{equation}
    A(k)
    =
    \frac{1}{N-k}
    \sum_{i=1}^{N-k} c_{(i)}.
\end{equation}
The AURAC is then defined as the empirical area under this rejection accuracy curve:
\begin{equation}
    \mathrm{AURAC}
    =
    \frac{1}{N}
    \sum_{k=0}^{N-1} A(k).
\end{equation}
A higher AURAC indicates that the estimator assigns high uncertainty to incorrect predictions, so that rejecting uncertain samples improves the accuracy of the retained subset.

\textbf{Balanced Accuracy}
We also report balanced accuracy for binary uncertainty detection. Since the uncertainty score $u_i$ is continuous, we first select a threshold $\tau^\ast$ on the validation set using the ROC curve. In practice, we choose the operating point that maximises Youden's index~\cite{ruopp2008youden}:
\begin{equation}
    \tau_\mathrm{you}^\ast
    =
    \arg\max_{\tau_\mathrm{you}}
    \left(
    \mathrm{TPR}_{\mathrm{val}}(\tau)
    -
    \mathrm{FPR}_{\mathrm{val}}(\tau)
    \right),
\end{equation}
where incorrect generations are treated as the positive class. The same threshold $\tau^\ast$ is then applied to the test set. A test example is predicted to be incorrect if its uncertainty score is above the threshold:
\begin{equation}
    \hat{w}_i
    =
    \mathbb{I}[u_i \geq \tau_\mathrm{you}^\ast],
\end{equation}
where $\hat{w}_i=1$ denotes a predicted unreliable generation. Let $w_i=1-c_i$ denote the ground-truth unreliability label. Balanced accuracy is then defined as the average of the true positive rate and true negative rate:
\begin{equation}
    \mathrm{Bal.\ Acc}
    =
    \frac{1}{2}
    \left(
    \frac{\sum_{i=1}^{N}\mathbb{I}[\hat{w}_i=1]\mathbb{I}[w_i=1]}
    {\sum_{i=1}^{N}\mathbb{I}[w_i=1]}
    +
    \frac{\sum_{i=1}^{N}\mathbb{I}[\hat{w}_i=0]\mathbb{I}[w_i=0]}
    {\sum_{i=1}^{N}\mathbb{I}[w_i=0]}
    \right),
\end{equation}
where $\mathbb{I}[\cdot]$ denotes the indicator function, which equals $1$ if the condition inside the brackets is true and $0$ otherwise. Balanced accuracy accounts for class imbalance cases.

\subsection{Evaluation Settings}
\label{app_subsec:evaluation settings}

We evaluate on three extractive question-answering benchmarks: CoQA, NewsQA, and emrQA. Each dataset is partitioned into training, validation, and test splits. When there is no predefined validation set, 10\% of the available training data is held out and used exclusively for hyperparameter selection and early stopping.

The MetaUE model consists of a frozen encoder followed by a two-layer MLP classifier with hidden dimension 256, SiLU activation, and a configurable dropout rate. Qwen3-VL-Embedding-2B~\cite{li2026qwen3} is used as the default encoder.
Models are trained using the AdamW optimiser with a linear warmup over the first 10\% of training steps, followed by cosine annealing decay. We perform a grid search over learning rate $\in \{3 \times 10^{-3},\, 1 \times 10^{-3},\, 3 \times 10^{-4}\}$ and batch size $\in \{64, 128, 256\}$, selecting the configuration that maximises validation AUROC. Training runs for up to 100 epochs with early stopping triggered if validation AUROC does not improve within 10\% of the total training steps. MetaUE is trained and evaluated over five independent runs with different seeds. 

All primary results are reported as mean $\pm$ standard deviation, which is estimated via 1000 bootstrap samples drawn from the test set predictions of each trained model, with the reported standard deviation reflecting variability across both seeds and bootstrap resamples. Evaluation metrics include AUROC, AURAC, and balanced accuracy computed on ground-truth correctness labels. Following the criterion in~\cite{kuhn2023semantic}, we treat a generated answer as correct if its ROUGE-L score against the reference answer exceeds $0.3$. The main experiments are conducted on an NVIDIA RTX 6000 Ada GPU with 48GB VRAM, while experiments involving Gemma4-26B, Gemma4-31B, and Qwen3.5-35B are conducted on an NVIDIA RTX 6000 Pro GPU with 96GB VRAM. All the computing resources have 256GB RAM and an Intel Xeon w7-2575X CPU.

\section{Notations}
\label{app:notation}

The notation used throughout the paper is illustrated in Table~\ref{tab:notation}.

\begin{table*}[h!]
\centering
\caption{Notations used throughout the paper.}
\label{tab:notation}
\small
\setlength{\tabcolsep}{4pt}
\renewcommand{\arraystretch}{1.15}
\begin{tabular}{p{0.18\textwidth} p{0.75\textwidth}}
\toprule
\textbf{Notation} & \textbf{Meaning} \\
\midrule
$\theta$ & Pretrained conditional generative model, i.e., the LLM. \\
$\mathbf{z}$ & Input prompt. \\
$\mathbf{z}_i$ & The $i$-th training prompt. \\
$\mathbf{Y}=(y_1,\ldots,y_T)$ & Generated response sequence. \\
$y_t$ & Token generated at step $t$. \\
$T$ & Full generation length. \\
$\mathcal{V}$ & Vocabulary of the LLM. \\
$p_{\theta}(\mathbf{y}\mid\mathbf{z})$ & Conditional probability of generating response $\mathbf{y}$ given prompt $\mathbf{z}$. \\
$p_{\theta}(y_t\mid\mathbf{z},\mathbf{y}_{<t})$ & Next-token distribution at generation step $t$. \\
$\mathbf{\ell}_t\in\mathbb{R}^{|\mathcal{V}|}$ & Logit vector over the vocabulary before softmax at step $t$. \\
$\ell_{t,v}$ & Logit assigned to candidate token $v$ at step $t$. \\
$s_t$ & Observable internal state at step $t$. In this work, $s_t$ is the logit vector $\mathbf{\ell}_t$. \\
$\mathbf{y}^{\star}$ & Reference answer or task-specific evaluation target. \\
$f(\mathbf{Y},\mathbf{y}^{\star})$ & Task-specific correctness function, with value in $[0,1]$. \\
$C$ & Correctness variable, $C=f(\mathbf{Y},\mathbf{y}^{\star})$. \\
$U$ & Unreliability variable, $U=1-C=1-f(\mathbf{Y},\mathbf{y}^{\star})$. \\
$\mathcal{I}$ & Information available to an uncertainty estimator. \\
$u_{\psi}(\mathcal{I})$ & Uncertainty estimator parameterised by $\psi$. Larger values indicate greater predicted unreliability. \\
$u^{\star}(\mathcal{I})$ & Ideal uncertainty estimator, $u^{\star}(\mathcal{I})=\mathbb{E}[U\mid\mathcal{I}]$. \\
$\mathcal{F}_0$ & Information available before generation, $\mathcal{F}_0=\sigma(\mathbf{z})$. \\
$\mathcal{F}_t$ & Information available after observing the first $t$ generated tokens and states. \\
$\mathcal{F}^{(R)}_T$ & Information from $R$ independently sampled full generations. \\
$R$ & Number of sampled generations used by multi-generation estimation. \\
$\mathcal{C}_{\mathrm{gen}}$ & Computational cost of one full autoregressive generation. \\
$N$ & Fixed prefix length used in partial-generation analysis or illustration. \\
$\tau$ & Adaptive stopping time for partial-generation uncertainty estimation, with $\tau\leq T$. \\
$\tau_i$ & Sample-dependent stopping time for the $i$-th prompt. \\
$\phi$ & Frozen pretrained text encoder used by MetaUE. \\
$g_{\omega}$ & Trainable two-layer MLP head in MetaUE. \\
$\omega$ & Trainable parameters of the MetaUE prediction head. \\
$f_{\omega}(\mathbf{z})$ & MetaUE prediction for prompt $\mathbf{z}$, defined as $f_{\omega}(\mathbf{z})=g_{\omega}(\phi(\mathbf{z}))$. \\
$n$ & Number of training prompts used to train MetaUE. \\
$U_{\tau_i}$ & Early-stopped Logit Magnitude score used as the pseudo-label for prompt $\mathbf{z}_i$. \\
$\mathcal{J}(\omega)$ & Mean-squared error training objective for MetaUE. \\
$X\in L^2$ & Scalar target to be estimated, such as final unreliability. \\
$M_t$ & Doob martingale estimate after observing $\mathcal{F}_t$, $M_t=\mathbb{E}[X\mid\mathcal{F}_t]$. \\
$\Delta_t$ & Martingale increment at step $t$, $\Delta_t=M_t-M_{t-1}$. \\
$\mathcal{E}(\tau)$ & Error penalty of stopping at $\tau$ rather than observing the full trajectory. \\
$\epsilon$ & Upper bound on the conditional update energy in the unobserved tail. \\
$\mathbf{1}_{\{t>\tau\}}$ & Indicator that token $t$ lies in the unobserved tail after stopping. \\
$\mathbb{E}[\cdot]$ & Expectation. \\
$\sigma(\cdot)$ & Sigma-algebra generated by the enclosed random variables. \\
\bottomrule
\end{tabular}
\end{table*}

\section{Hyperparameter Sensitivity of the Logit Magnitude}
\label{sec:mw_logit_magnitude}

\begin{figure*}[t]
  \centering

  \begin{subfigure}[t]{0.325\linewidth}
    \includegraphics[width=\linewidth]{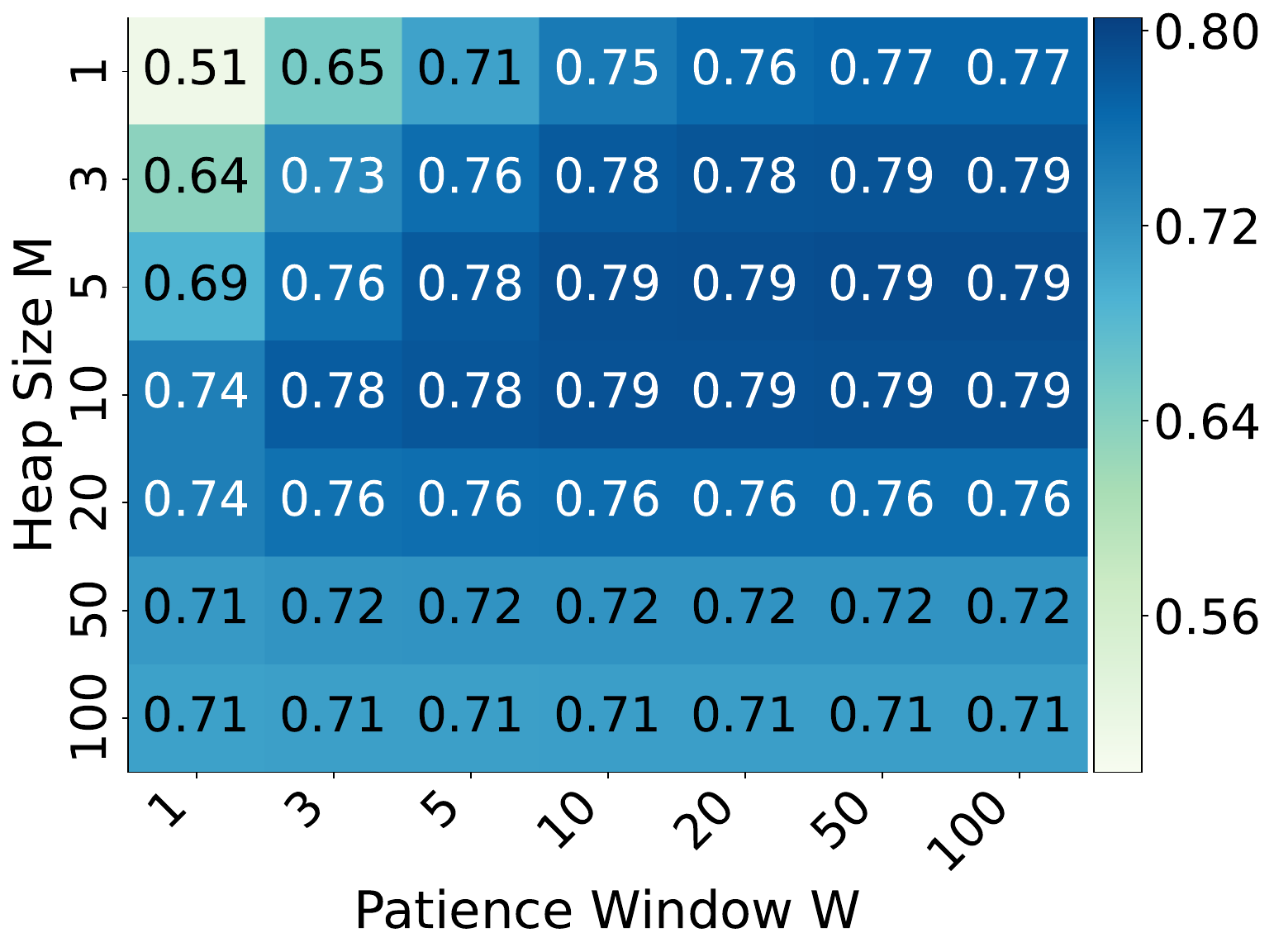}
    \caption{CoQA, Qwen (AUROC)}
  \end{subfigure}\hfill
  \begin{subfigure}[t]{0.325\linewidth}
    \includegraphics[width=\linewidth]{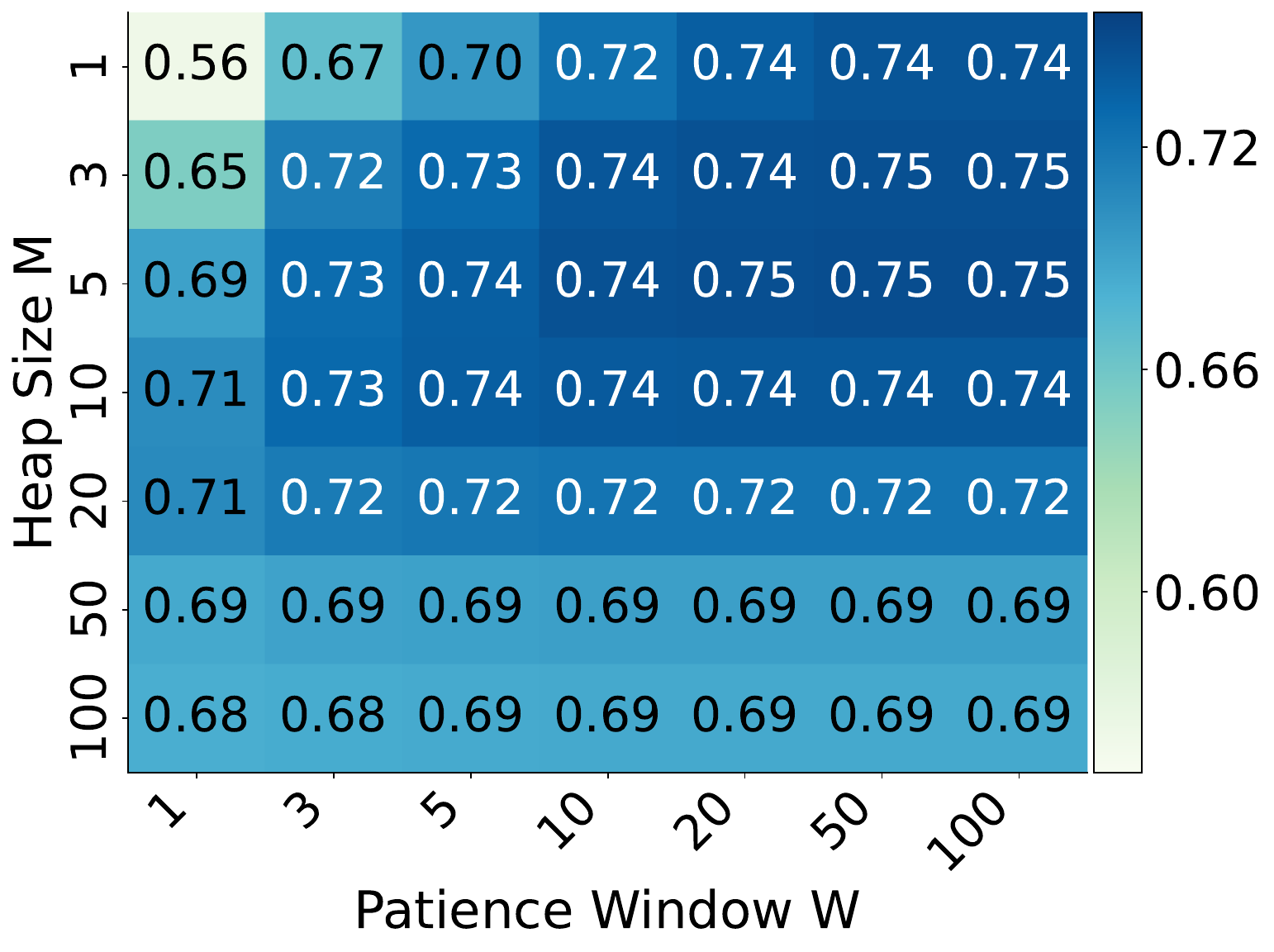}
    \caption{NewsQA, Qwen (AUROC)}
  \end{subfigure}\hfill
  \begin{subfigure}[t]{0.325\linewidth}
\includegraphics[width=\linewidth]{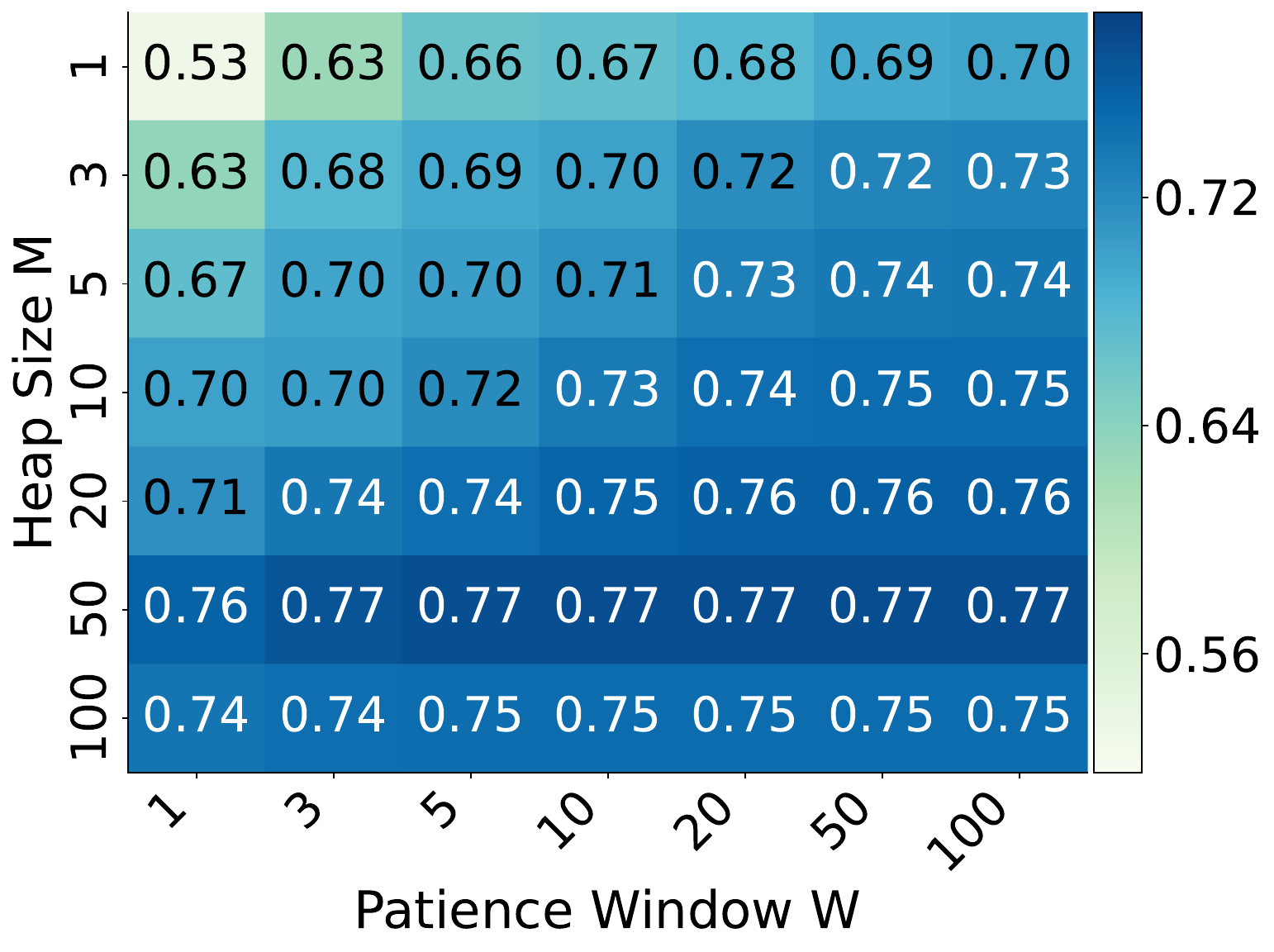}
    \caption{EmrQA, Qwen (AUROC)}
  \end{subfigure}
  \vspace{0.5em}

  \begin{subfigure}[t]{0.325\linewidth}
    \includegraphics[width=\linewidth]{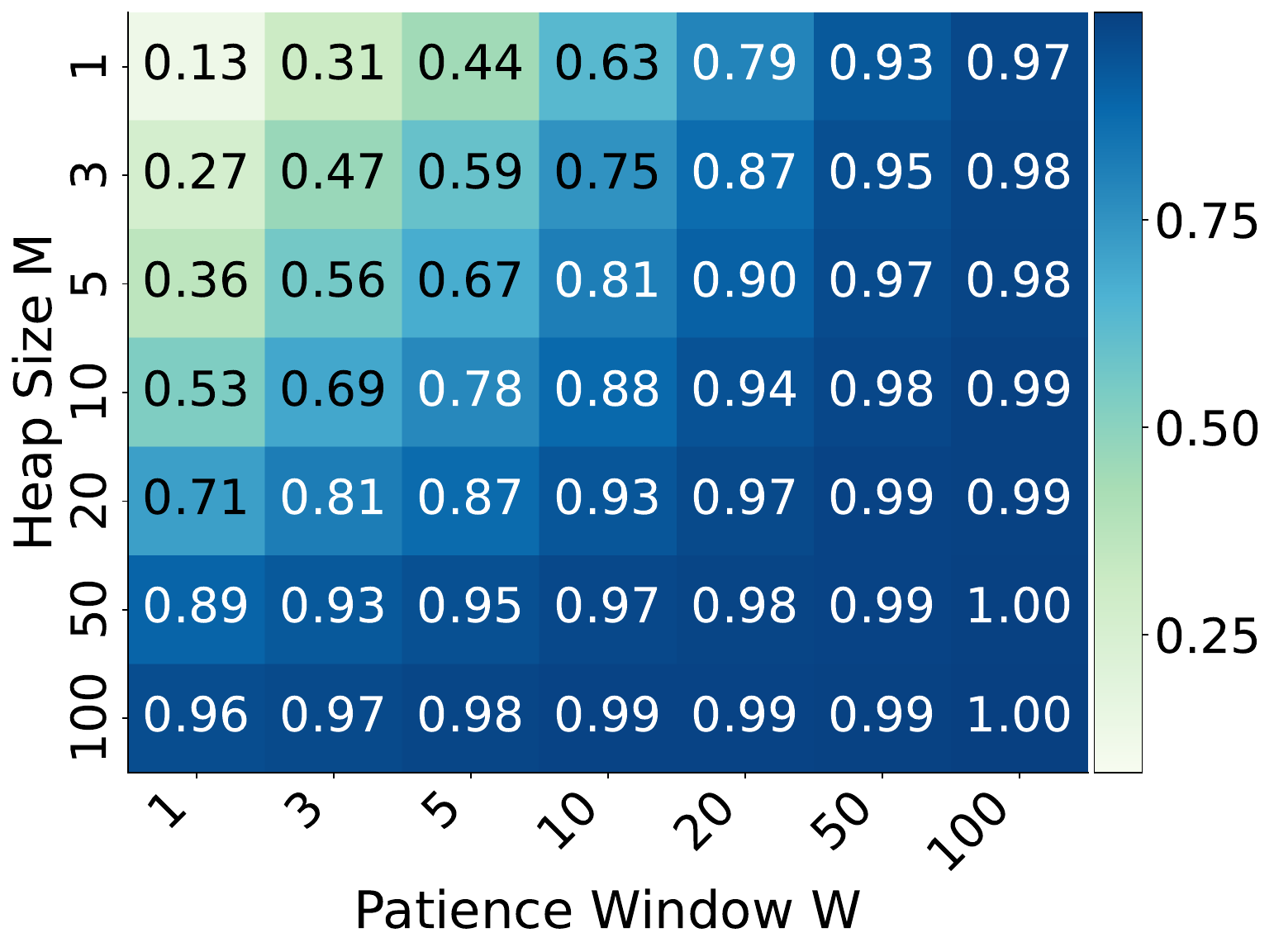}
    \caption{CoQA, Qwen ($\tau/T$)}
  \end{subfigure}\hfill
  \begin{subfigure}[t]{0.325\linewidth}
    \includegraphics[width=\linewidth]{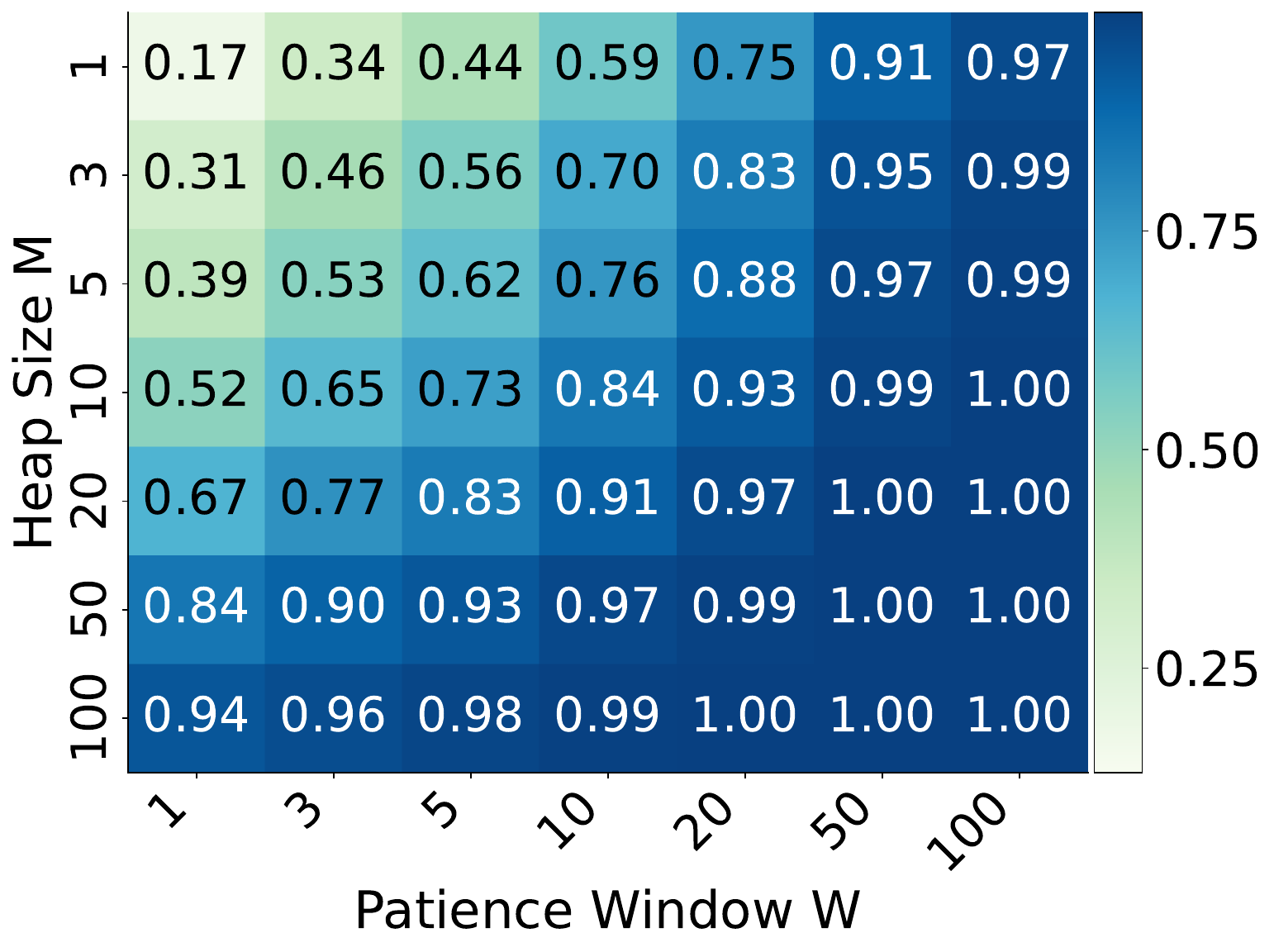}
    \caption{NewsQA, Qwen ($\tau/T$)}
  \end{subfigure}\hfill
  \begin{subfigure}[t]{0.325\linewidth}
    \includegraphics[width=\linewidth]{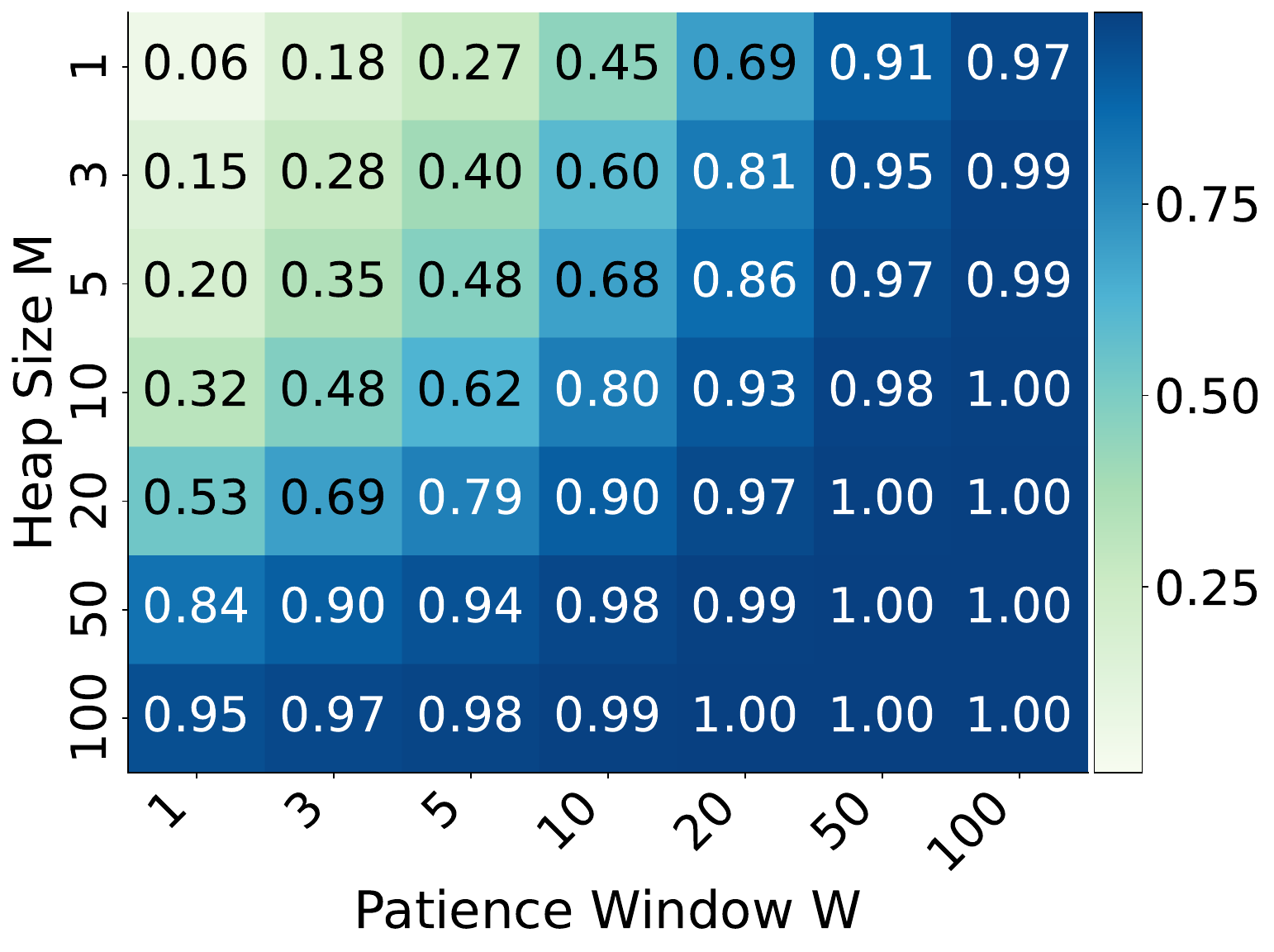}
    \caption{EmrQA, Qwen ($\tau/T$)}
  \end{subfigure}

  \caption{Plots (a)-(c) shows heatmaps of AUROC and Plots (d)-(f) shows mean
    token-consumption ratio ($\tau/T$) on the validation set.
    Each cell corresponds to a patience-rule configuration
    $(M, W)$, with $M$ (top-$M$ for token-level score aggregation) on the vertical axis and
    $W$ (patience window) on the horizontal axis.}
  \label{fig:patience_heatmaps}
\end{figure*}

We evaluate how the top-$M$ aggregation and the patience window $W$ of the patience rule affect the logit-magnitude uncertainty estimation. We sweep $M \in \{1, 3, 5, 10, 20, 50, 100\}$ and $W \in \{1, 3, 5, 10, 20, 50, 100\}$ on the validation splits of CoQA, NewsQA, and EmrQA for Qwen3.5-4B, reporting AUROC as the performance metric and the token-consumption ratio $\tau/T$ as the efficiency metric. 
Figure~\ref{fig:patience_heatmaps} reveals that the optimal $(M, W)$ configuration varies across datasets.
On CoQA and NewsQA, where mean generation lengths are modest (${\sim}17$--$21$ tokens), an $M = 5$ achieves near-peak AUROC across both models with windows $W \geq 10$. On EmrQA, where Qwen3.5-4B generates substantially longer sequences (${\sim}42$ tokens on average), larger $M = 50$ are needed to capture a sufficient uncertainty signal. While calibration on held-out validation data is desirable in practice, when no data is available, $M \in \{5, 10\}$ provides a robust starting point, with $W \approx 2M$ offering a reasonable patience window.

\section{Transferability of MetaUE}
\label{append_sec:transfer}

\begin{table}[h]
  \centering
  \caption{%
    Dataset transferability of MetaUE. Diagonal entries are in-domain results included as a reference upper bound. Each metric is reported in the mean$_\text{std}$ format.}
  \label{tab:dataset_transfer}

  \begin{subtable}{\linewidth}
    \centering
    \subcaption{Gemma4-4B}
    \label{tab:dataset_transfer_gemma}
    \setlength{\tabcolsep}{5pt}
    \small
    \begin{tabular}{l *{6}{c}}
      \toprule
     Target $\rightarrow$ & \multicolumn{2}{c}{CoQA}
      & \multicolumn{2}{c}{NewsQA}
      & \multicolumn{2}{c}{emrQA} \\
      \cmidrule(lr){2-3}\cmidrule(lr){4-5}\cmidrule(lr){6-7}
      Source $\downarrow$ & AUROC & AURAC & AUROC & AURAC & AUROC & AURAC \\
      \midrule
      CoQA
        & $0.709_{.004}$ & $0.475_{.003}$
        & $0.613_{.002}$ & $0.592_{.002}$
        & $0.598_{.003}$ & $0.678_{.003}$ \\[2pt]
      NewsQA
        & $0.647_{.014}$ & $0.439_{.009}$
        & $0.617_{.003}$ & $0.599_{.002}$
        & $0.580_{.011}$ & $0.661_{.008}$ \\[2pt]
      emrQA
        & $0.567_{.004}$ & $0.392_{.003}$
        & $0.563_{.005}$ & $0.559_{.003}$
        & $0.719_{.003}$ & $0.742_{.002}$ \\
      \bottomrule
    \end{tabular}
  \end{subtable}

  \begin{subtable}{\linewidth}
    \centering
    \subcaption{Qwen3.5-4B}
    \label{tab:dataset_transfer_qwen}
    \setlength{\tabcolsep}{5pt}
    \small
    \begin{tabular}{l *{6}{c}}
      \toprule
      Target $\rightarrow$& \multicolumn{2}{c}{CoQA}
      & \multicolumn{2}{c}{NewsQA}
      & \multicolumn{2}{c}{emrQA} \\
      \cmidrule(lr){2-3}\cmidrule(lr){4-5}\cmidrule(lr){6-7}
      Source $\downarrow$ & AUROC & AURAC & AUROC & AURAC & AUROC & AURAC \\
      \midrule
      CoQA
        & $0.724_{.003}$ & $0.564_{.001}$
        & $0.715_{.004}$ & $0.789_{.002}$
        & $0.618_{.006}$ & $0.518_{.004}$ \\[2pt]
      NewsQA
        & $0.719_{.008}$ & $0.555_{.006}$
        & $0.717_{.002}$ & $0.788_{.001}$
        & $0.545_{.004}$ & $0.480_{.004}$ \\[2pt]
      emrQA
        & $0.584_{.008}$ & $0.486_{.005}$
        & $0.636_{.007}$ & $0.750_{.004}$
        & $0.733_{.004}$ & $0.588_{.002}$ \\
      \bottomrule
    \end{tabular}
  \end{subtable}
\end{table}

\begin{table}[h]
  \centering
  \caption{%
    Model transferability of MetaUE.
    Rows are grouped by target LLM with the in-domain baseline and the target model. Each metric is reported in the mean$_\text{std}$ format. \emph{Gemma} denotes Gemma4-4B; \emph{Qwen} denotes Qwen3.5-4B.
  }
  \label{tab:model_transfer}
  \vspace{4pt}
  \setlength{\tabcolsep}{5pt}
  \small
  \begin{tabular}{l *{6}{c}}
    \toprule
    & \multicolumn{2}{c}{CoQA}
    & \multicolumn{2}{c}{NewsQA}
    & \multicolumn{2}{c}{emrQA} \\
    \cmidrule(lr){2-3}\cmidrule(lr){4-5}\cmidrule(lr){6-7}
    & AUROC & AURAC & AUROC & AURAC & AUROC & AURAC \\
    \midrule
    \quad Qwen (in-domain)
      & $0.724_{.003}$ & $0.564_{.001}$
      & $0.717_{.002}$ & $0.788_{.001}$
      & $0.733_{.004}$ & $0.588_{.002}$ \\[2pt]
    \quad Gemma $\to$ Qwen
      & $0.719_{.003}$ & $0.566_{.002}$
      & $0.705_{.004}$ & $0.784_{.002}$
      & $0.723_{.003}$ & $0.584_{.003}$ \\
    \midrule
    \quad Gemma (in-domain)
      & $0.709_{.004}$ & $0.475_{.003}$
      & $0.617_{.003}$ & $0.599_{.002}$
      & $0.719_{.003}$ & $0.742_{.002}$ \\[2pt]
    \quad Qwen $\to$ Gemma
      & $0.675_{.004}$ & $0.450_{.002}$
      & $0.576_{.004}$ & $0.574_{.003}$
      & $0.657_{.004}$ & $0.710_{.002}$ \\
    \bottomrule
  \end{tabular}
\end{table}

In this experiment, we evaluate the transferability of MetaUE under two conditions:
(i) dataset transferability, where a model trained on one
dataset is applied to a different dataset using the same LLM; and
(ii) model transferability, where a model trained on one LLM is applied to a different LLM on the same dataset. All configurations use the logit-magnitude label function with Qwen3-VL-Embedding-2B embeddings and are evaluated with AUROC and AURAC.

MetaUE shows strong transferability in several settings, with cross-condition performance often approaching the in-domain baseline. For dataset transfer (Table~\ref{tab:dataset_transfer}), transfer between the two general-domain datasets, COQA and NewsQA, is particularly stable. For example, on Qwen3.5-4B, COQA$\to$NewsQA achieves an AUROC of $0.715$, compared with $0.717$ for in-domain training. In contrast, transfer involving emrQA leads to larger degradation, reflecting the stronger domain shift introduced by clinical text. For instance, on Gemma4-4B, transferring from emrQA to COQA reduces AURAC to $0.392$. For model transfer (Table~\ref{tab:model_transfer}), Gemma4-4B$\to$Qwen3.5-4B is highly effective, achieving AUROC within $0.01$ of Qwen's in-domain baseline across all datasets ($0.705$-$0.723$ versus $0.717$-$0.733$). By comparison, Qwen3.5-4B$\to$Gemma4-4B shows a larger gap ($0.576$-$0.675$ versus $0.617$-$0.719$), suggesting that transferability is asymmetric and depends on how well the source model's uncertainty behaviour matches that of the target model.

\section{Evaluation of Logit Magnitude and MetaUE on Llama3}
\label{app_sec:evaluation on llama3}

\begin{table*}[h]
\centering
\caption{
Assessment of uncertainty estimation methods on Llama3. \textbf{Bold} values indicate the best performance.
}
\footnotesize
\setlength{\tabcolsep}{2.0pt}
\renewcommand{\arraystretch}{1.08}
\label{table:llama3}
\resizebox{\textwidth}{!}{
\begin{tabular}{lll|cccc|cccc|cccc}
\toprule
\multirow{2}{*}{\textbf{LLM}}
& \multirow{2}{*}{\textbf{Size}}
& \multirow{2}{*}{\textbf{Method}}
& \multicolumn{4}{c|}{\textbf{COQA}}
& \multicolumn{4}{c|}{\textbf{NewsQA}}
& \multicolumn{4}{c}{\textbf{emrQA}} \\
\cmidrule(lr){4-7} \cmidrule(lr){8-11} \cmidrule(lr){12-15}
& & & AUROC $\uparrow$ & AURAC $\uparrow$ & Bal. Acc $\uparrow$ & N-tok $\downarrow$
  & AUROC $\uparrow$ & AURAC $\uparrow$ & Bal. Acc $\uparrow$ & N-tok $\downarrow$
  & AUROC $\uparrow$ & AURAC $\uparrow$ & Bal. Acc $\uparrow$ & N-tok $\downarrow$ \\
\midrule

\multirow{8}{*}{\rotatebox[origin=c]{90}{Llama3}}
& \multirow{4}{*}{3B}
& LogTokU
    & \textbf{0.751}$_{.005}$ & \textbf{0.499}$_{.007}$ & 0.675$_{.005}$ & 18.9
    & \textbf{0.728}$_{.008}$ & \textbf{0.532}$_{.010}$ & 0.666$_{.007}$ & 29.0
    & 0.616$_{.003}$ & 0.518$_{.003}$ & 0.600$_{.003}$ & 27.4 \\
& & Self-Certainty
    & 0.703$_{.006}$ & 0.471$_{.007}$ & 0.647$_{.006}$ & 18.9
    & 0.603$_{.009}$ & 0.452$_{.010}$ & 0.572$_{.007}$ & 29.0
    & \textbf{0.725}$_{.003}$ & \textbf{0.600}$_{.003}$ & \textbf{0.666}$_{.003}$ & 27.4 \\
\cmidrule{3-15}
& & Logit Magnitude
    & 0.731$_{.006}$ & 0.482$_{.007}$ & \textbf{0.677}$_{.005}$ & 18.9
    & 0.727$_{.008}$ & 0.530$_{.010}$ & \textbf{0.673}$_{.007}$ & 29.0
    & 0.655$_{.003}$ & 0.552$_{.003}$ & 0.618$_{.003}$ & 27.4 \\
& & MetaUE
    & 0.656$_{.006}$ & 0.430$_{.004}$ & 0.624$_{.005}$ & 0
    & 0.577$_{.002}$ & 0.438$_{.002}$ & 0.551$_{.004}$ & 0
    & 0.609$_{.004}$ & 0.531$_{.003}$ & 0.577$_{.001}$ & 0 \\
\cmidrule(lr){2-15}
& \multirow{4}{*}{8B}
& LogTokU
    & \textbf{0.773}$_{.005}$ & \textbf{0.516}$_{.007}$ & 0.696$_{.005}$ & 23.1
    & \textbf{0.757}$_{.007}$ & 0.633$_{.008}$ & 0.683$_{.007}$ & 27.2
    & 0.631$_{.003}$ & 0.518$_{.003}$ & 0.607$_{.003}$ & 20.4 \\
& & Self-Certainty
    & 0.717$_{.006}$ & 0.474$_{.007}$ & 0.658$_{.005}$ & 23.1
    & 0.634$_{.009}$ & 0.562$_{.010}$ & 0.590$_{.008}$ & 27.2
    & \textbf{0.742}$_{.003}$ & \textbf{0.602}$_{.003}$ & \textbf{0.681}$_{.003}$ & 20.4 \\
\cmidrule{3-15}
& & Logit Magnitude
    & 0.756$_{.005}$ & 0.504$_{.007}$ & \textbf{0.703}$_{.005}$ & 23.1
    & 0.739$_{.008}$ & \textbf{0.634}$_{.007}$ & \textbf{0.687}$_{.007}$ & 23.1
    & 0.651$_{.003}$ & 0.548$_{.003}$ & 0.612$_{.003}$ & 20.4 \\
& & MetaUE
    & 0.679$_{.005}$ & 0.453$_{.003}$ & 0.635$_{.005}$ & 0
    & 0.582$_{.002}$ & 0.529$_{.002}$ & $0.557_{.002}$ & 0
    & 0.643$_{.004}$ & 0.541$_{.003}$ & 0.602$_{.003}$ & 0 \\
\bottomrule
\end{tabular}}
\end{table*}

\begin{figure*}[h!]
  \centering
  \begin{subfigure}[t]{0.325\linewidth}
    \includegraphics[width=\linewidth]{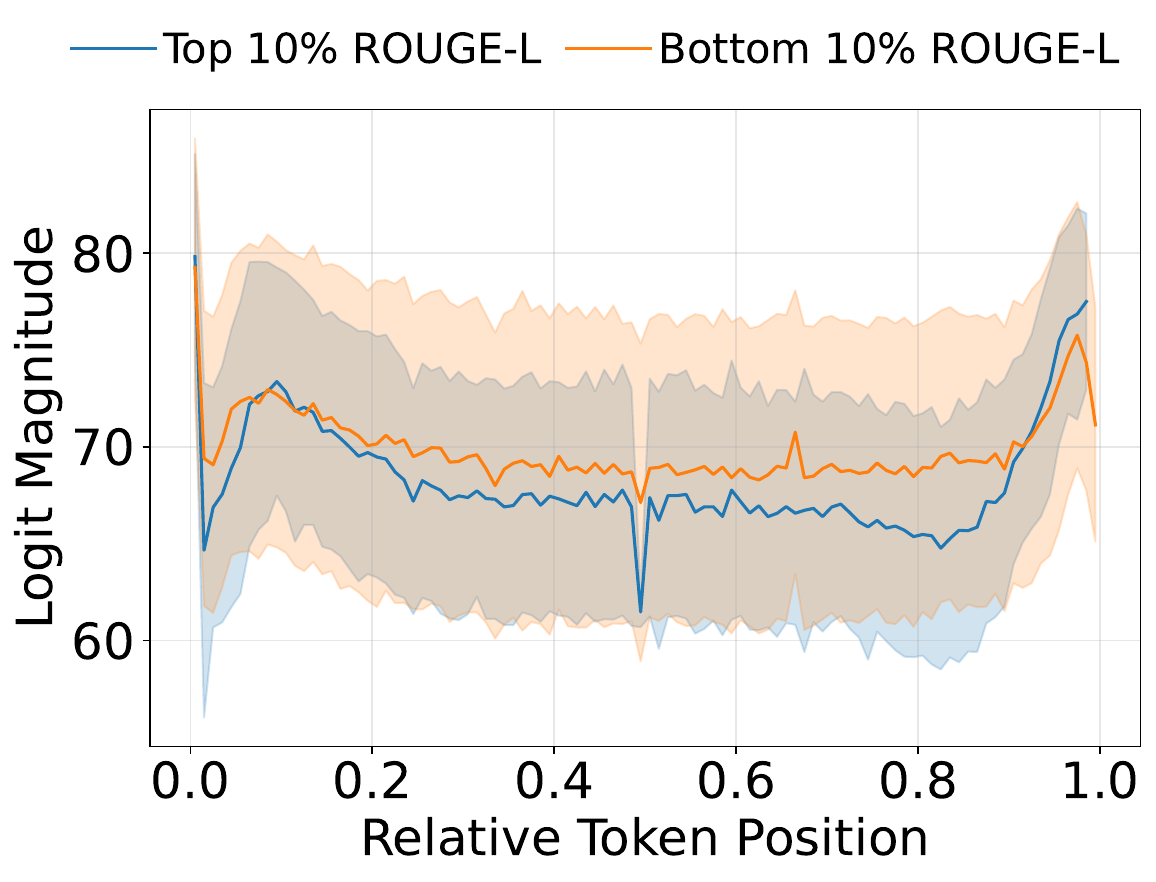}
    \caption{Llama3-3B}
    \label{fig:emrqa_mag_llama}
  \end{subfigure}\hfill
  \begin{subfigure}[t]{0.325\linewidth}
    \includegraphics[width=\linewidth]{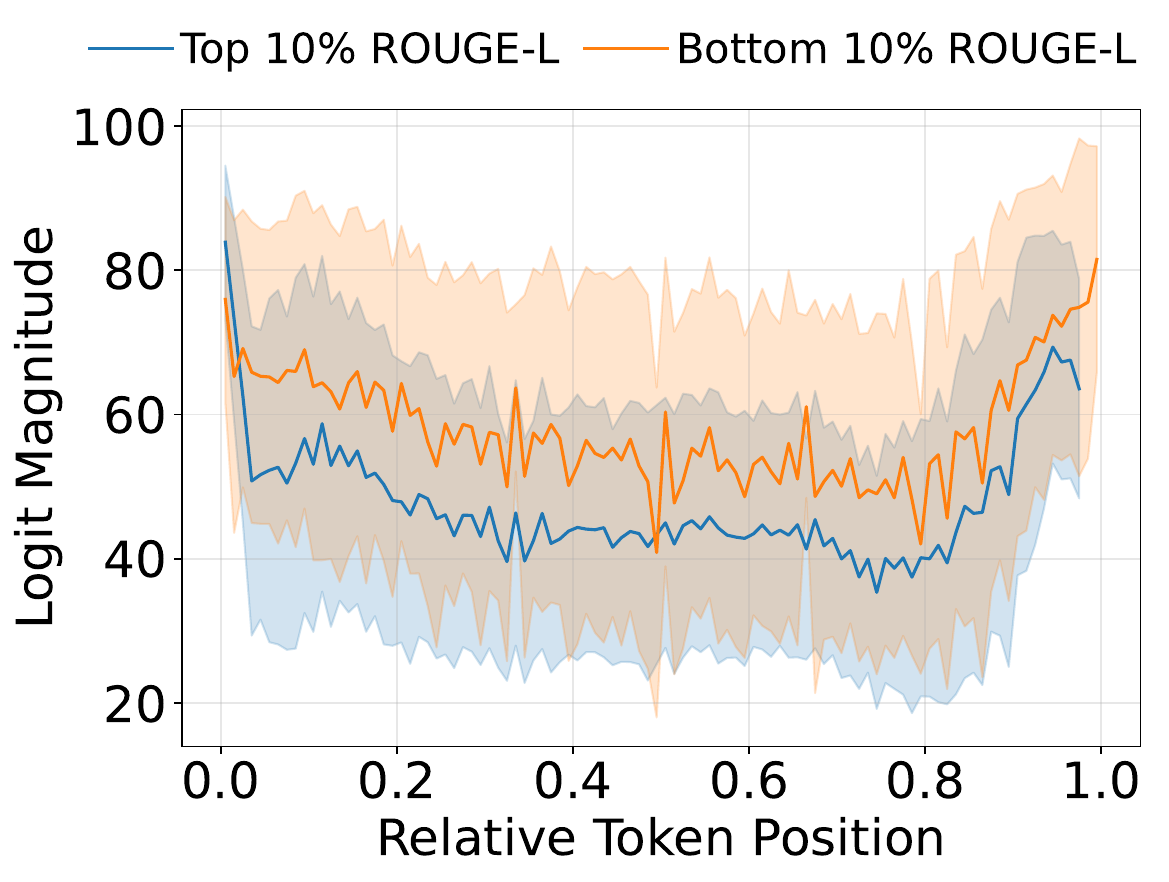}
    \caption{Gemma4-4B}
    \label{fig:emrqa_mag_gemma}
  \end{subfigure}\hfill
  \begin{subfigure}[t]{0.325\linewidth}
    \includegraphics[width=\linewidth]{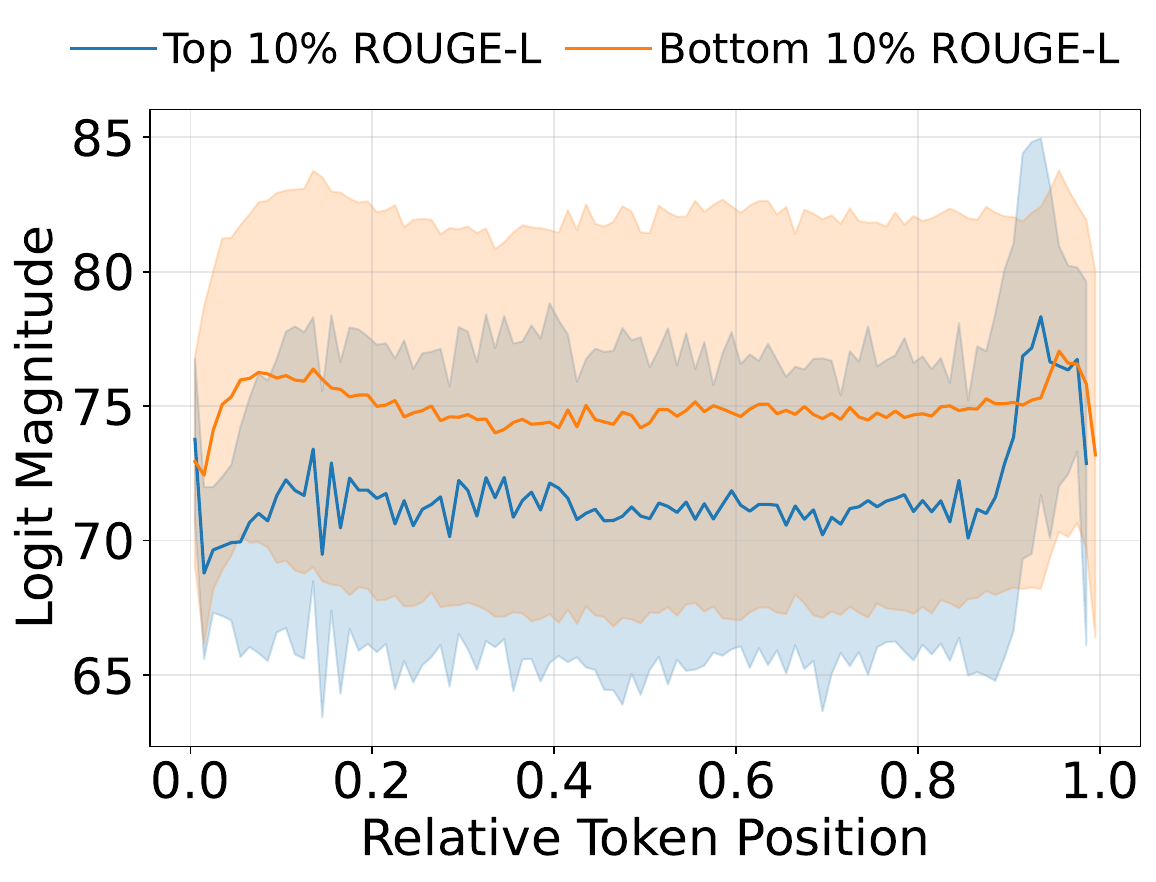}
    \caption{Qwen3.5-4B}
    \label{fig:emrqa_mag_qwen}
  \end{subfigure}
  \caption{Token-level Logit Magnitude (mean\,$\pm$\,std) at different relative token positions on emrQA. A larger separation between the two curves indicates a stronger discriminative signal.}
  \label{fig:logit_magnitude_emrqa}
\end{figure*}

As shown in Table~\ref{table:llama3}, Logit Magnitude underperforms Self-Certainty with Llama3 on emrQA. A potential reason is that RLHF-trained Llama3 models exhibit sharpened probability distributions on both correct and incorrect answers in the medical context~\citep{bentegeac2025token}. This sharpened distribution makes logit norms saturate and lose the capabilities in estimating uncertainty. 
Figure~\ref{fig:logit_magnitude_emrqa} shows per-token Logit Magnitude for the top and bottom 10\% of emrQA samples ordered by ROUGE-L, which is used to determine the correctness~\cite{kuhn2023semantic}. For Gemma and Qwen, the bottom-10\% curve lies above the top-10\%, providing a clear discriminative signal.
For Llama3-3B, the two curves nearly coincide with overlapping standard-deviation bands, leaving the logit magnitude with little discriminative signal. The plot shows the severe logit saturation caused by RLHF~\citep{bentegeac2025token}. Future work will explore whether applying post-hoc temperature scaling to soften the logit-level distributions can restore the gap between these curves and recover the uncertainty signal.

MetaUE degrades on Llama3, most notably on NewsQA. This drop is not simply inherited from a weaker teacher, since Logit Magnitude itself remains competitive there. We hypothesise two factors. (i) the frozen Qwen3-VL encoder may emphasise semantic features aligned with the latest LLMs rather than the Llama3; and (ii) NewsQA's long passages induce generation-based uncertainty that the prompt embedding cannot recover.

\section*{Broader Impacts}
\label{app_sec:impact}

This work contributes to more efficient assessment for LLM generation by reducing the amount of generation needed for uncertainty estimation. Earlier uncertainty signals may help users identify potentially unreliable outputs before they are acted upon, which is especially valuable in latency-sensitive or high-stakes applications such as clinical decision support. 
More broadly, the proposed methods could support selective prediction, human-in-the-loop review, and resource-aware deployment by prioritising which generations require additional verification. At the same time, uncertainty scores should be interpreted as decision-support signals rather than guarantees of correctness. Responsible use requires clear communication of uncertainty, appropriate domain-specific calibration, and integration with human oversight when decisions may affect safety, health, or access to services.



\end{document}